%% file: main.tex
\documentclass[runningheads]{llncs}

\usepackage{eccv}

\usepackage{eccvabbrv}
\usepackage{bm}
\usepackage{algorithm}
\usepackage[noend]{algpseudocode}
\usepackage{graphicx}
\usepackage{booktabs}
\usepackage[accsupp]{axessibility}  % Improves PDF readability for those with disabilities.
\usepackage{caption}
\usepackage{amsmath}
\usepackage{algpseudocode}
\usepackage{multirow}
\usepackage{wrapfig}
\usepackage{makecell}
\usepackage{colortbl}  %彩色表格需要加载的宏包
\usepackage{xcolor}
\usepackage{array}   %对表列和表格线的设置需要用到array宏包
\usepackage{hyperref}

\usepackage{orcidlink}

\begin{document}

\title{DreamDrone: Text-to-Image Diffusion Models\\are Zero-shot Perpetual View Generators} 

\titlerunning{DreamDrone}

\author{Hanyang Kong\inst{1}\orcidlink{0000-0002-5895-5112} \and
Dongze Lian\inst{1}\orcidlink{0000-0002-4947-0316} \and
Michael Bi Mi\inst{2}\orcidlink{0009-0000-4930-1849} \and
Xinchao Wang\inst{1}\thanks{Corresponding author.}\orcidlink{0000-0003-0057-1404}}

\authorrunning{H.~Kong et al.}

\institute{National University of Singapore, Singapore\\
\and
Huawei International Pte. Ltd., Singapore\\
\email{hanyang.k@u.nus.edu}, \email{dzlianx@gmail.com}, \email{xinchao@nus.edu.sg}}

%\maketitle

%\twocolumn[
{%
\renewcommand\twocolumn[1][]{#1}%
\maketitle
% \vspace{-0em}
\begin{center}
    \centering
    \captionsetup{type=figure}
    \includegraphics[width=1.0  \textwidth]{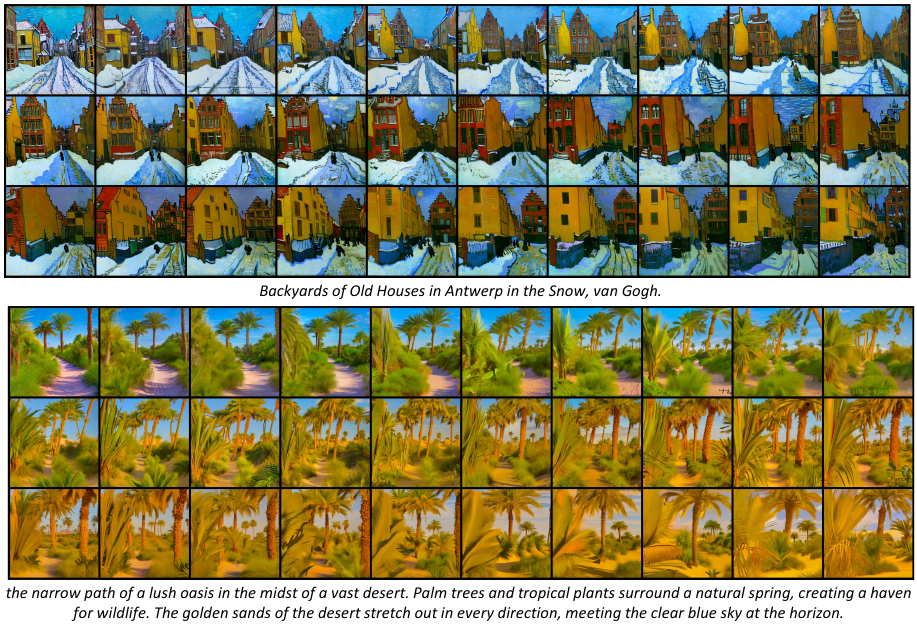}
    \captionof{figure}{\textbf{Visualization results of DreamDrone}. Given a single scene image and the textual description, our approach generates novel views corresponding to user-defined camera trajectory, without fine-tuning on any dataset or reconstructing the 3D point cloud in advance.}
    \label{fig1}
    % \vspace{-0.1em}
\end{center}%
}
%]

\input{sec/0_abstract}    
\input{sec/1_intro}

\input{sec/2_related_works}
\input{sec/3_method}
\input{sec/4_experiments}
\input{sec/5_conclusion}

% \title{DreamDrone: Text-to-Image Diffusion Models\\are Zero-shot Perpetual View Generators} 
% \maketitle
\newpage

\title{DreamDrone: Text-to-Image Diffusion Models\\are Zero-shot Perpetual View Generators\\
--- \emph{Supplement Material} --- } % TODO REVIEW: If the paper title is too long for the running head, you can set
\titlerunning{DreamDrone}

\author{Hanyang Kong\inst{1}\orcidlink{0000-0002-5895-5112} \and
Dongze Lian\inst{1}\orcidlink{0000-0002-4947-0316} \and
Michael Bi Mi\inst{2}\orcidlink{0009-0000-4930-1849} \and
Xinchao Wang\inst{1}\thanks{Corresponding author.}\orcidlink{0000-0003-0057-1404}}

\authorrunning{H.~Kong et al.}

\institute{National University of Singapore, Singapore\\
\and
Huawei International Pte. Ltd., Singapore\\
\email{hanyang.k@u.nus.edu}, \email{dzlianx@gmail.com}, \email{xinchao@nus.edu.sg}}
\maketitle
\input{sec/X_suppl}


{
    \small
    \bibliographystyle{splncs04}
    \bibliography{main}
}
\end{document}

%% file: sec/0_abstract.tex
\begin{abstract}

We introduce \textit{DreamDrone}, a novel zero-shot and training-free pipeline for generating unbounded flythrough scenes from textual prompts. Different from other methods that focus on warping images frame by frame, we advocate explicitly warping the intermediate latent code of the pre-trained text-to-image diffusion model for high-quality image generation and generalization ability.
To further enhance the fidelity of the generated images, we also propose a feature-correspondence-guidance diffusion process and a high-pass filtering strategy to promote geometric consistency and high-frequency detail consistency, respectively.
Extensive experiments reveal that \textit{DreamDrone} significantly surpasses existing methods, delivering highly authentic scene generation with exceptional visual quality, without training or fine-tuning on datasets or reconstructing 3D point clouds in advance.

%Leveraging this strategy, we further propose a feature-correspondence-guidance diffusion process, enabling the generation of novel views with geometry consistency. 
%Moreover, we propose a novel high-pass filtering strategy to preserve the consistency of high-frequency details across adjacent views. 

% a novel feature-correspondence-guidance diffusion process, which utilizes the strong correspondence of intermediate features in the diffusion model. Leveraging this guidance strategy, we further propose an advanced technique for editing the intermediate latent code, enabling the generation of subsequent novel views with geometric consistency. Extensive experiments reveal that \textit{DreamDrone} significantly surpasses existing methods, delivering highly authentic scene generation with exceptional visual quality. This approach marks a significant step in zero-shot perpetual view generation from textual prompts, enabling the creation of diverse scenes, including natural landscapes like oases and caves, as well as complex urban settings such as Lego-style street views.
\end{abstract}

%% file: sec/1_intro.tex
\section{Introduction}
\label{sec:intro}

Recent advances in vision and graphics have enabled the synthesis of multi-view consistent 3D scenes along extended camera trajectories
\cite{li2022infinitenature, liu2021infinite, cai2023diffdreamer, fridman2023scenescape}. This emerging task, termed \textit{perpetual view generation} \cite{liu2021infinite}, involves synthesizing views from a flying camera along an arbitrarily long trajectory, starting from a single RGBD image. 

Previous methodologies predominantly engage in warping images frame by frame with traditional 3D geometric knowledge when given RGBD images and subsequent camera extrinsic. 
However, this operation often leads to blurriness and distortion in images, which arises from inaccurate interpolation, the mismatch between discrete pixels and continuous transformations, and inaccurate depth data. Moreover, such blurriness and distortion tend to amplify with the accumulation of warp operations.

To further alleviate the errors caused by frame-by-frame warp operations, two primary paths have been proposed. i) Some methods~\cite{li2022infinitenature,liu2021infinite,cai2023diffdreamer} try to train a refiner on natural scene datasets. The advantage of this frame-by-frame approach is that it allows for arbitrary changes in camera trajectory during the scene generation process, offering users a higher degree of freedom and enabling infinite generation. However, this training-based method can only be used in natural scenes and cannot be generalized to arbitrary indoor/outdoor scenes or scenes of various styles. ii) Another solution is to first reconstruct the 3D scene model using text prompts, then render 2D RGB images according to the camera trajectory~\cite{fridman2023scenescape,yu2023wonderjourney,hollein2023text2room}. Although this solution yields more coherent 2D image sequences, the quality of the rendered images highly depends on the quality of the 3D scene model. This method cannot guarantee good rendering effects from every viewpoint. Additionally, since this method requires the reconstruction of 3D point clouds, it cannot achieve "infinite" scene generation in the same way as the frame-by-frame strategy.

In this paper, we advocate that a more general and flexible perpetual view generation pipeline should possess the following capabilities:

i) are versatile across diverse scenes, including indoor and outdoor scenes, as well as scenes depicted in various styles; ii) allow users to interactively control the camera trajectory during the process of scene generation, while ensuring the high quality of the generated images and the semantic consistency between adjacent frames; and iii) enable seamless transitions from one scene to another.

To this end, we introduce DreamDrone, a novel zero-shot, training-free, infinite scene generation pipeline from text prompts, which does not require any optimization or fine-tuning on any dataset. A core principle of our approach is to warp the latent code of a pre-trained text-to-image diffusion model rather than the frames, %to serve as a refiner, 
enriching it with temporal and geometric consistency. To be specific, given RGBD image $I$ of the current view and camera rotation $\mathcal{R}$ and translation $\mathcal{T}$ for the next view (which is interactively defined by users), we first obtain the latent code $x_{t_1}$ of the diffusion model at timestep $t_1$, warp it to latent code of the next view $x'_{t_1}$ based on $\mathcal{R}$ and $\mathcal{T}$, and denoise $x'_{t_1}$ to the image $I'$ of the next view. To ensure geometry consistency across adjacent views, we propose a novel feature-correspondence-guidance diffusion process when denoising from $x'_{t_1}$ to image at the next view $I'$. Moreover, we propose a novel high-pass filter mechanism when warping the latent code $x_{t_1}$, for preserving high-frequency details across adjacent views.

Our experiments demonstrate that the proposed DreamDrone effectively leads to high-quality and geometry-consistent scene generation. Quantitative and qualitative results demonstrate our comparable, even superior performance compared with other training-based and training-free methods from the aspects of temporal consistency and image quality. Moreover, the significant advantage of DreamDrone is its versatility: it is adept not only at generating real-world scenarios but also shows promising capabilities in creating imaginative scenes. Additionally, users can interactively control the camera trajectory (\cref{fig:camera_trajectory}) and shuttle from one scene to another (\cref{fig:change_prompt}). Our contributions are summarized as follows:
\begin{itemize}
    \item To our best knowledge, we are the first attempt to generate novel views by explicitly warping the latent code of the pre-trained diffusion model.
    \item A novel feature-correspondence-guidance diffusion process is proposed to enforce geometry consistency across adjacent views. Moreover, a high-pass filtering strategy is introduced to preserve high-frequency details for novel views.
    \item Extensive experiments demonstrate that our method generates high-quality and geometry-consistent novel views for any scene, from realistic to fantastical. More interestingly, our method realizes the scene shuttle, \textit{i.e.}, travels from one scene to another when the user controls the camera trajectory.
\end{itemize}

%% file: sec/2_related_works.tex
\section{Related Works}
\label{sec:related_works}

% inf-nat inf-nat-0, dreamdiff, text-to-room
% \vspace{-0.2cm}
\paragraph{Perpetual view generation.}
Perpetual view generation extrapolates unseen content outside a single image.
InfNat~\cite{liu2021infinite}, InfNat-0~\cite{li2022infinitenature}, and DiffDreamer~\cite{cai2023diffdreamer} use iterative training for long-trajectory perpetual view extrapolation. InfNat~\cite{liu2021infinite} pioneered the \textit{perpetual view generation} task with a database for infinite 2D landscapes. InfNat-0~\cite{li2022infinitenature} adapted this to 3D, introducing a render-refine-repeat phase for novel views. DiffDreamer~\cite{cai2023diffdreamer} improved consistency with image-conditioned diffusion models. However, these methods lack robustness in complex and urban environments.
In very recent concurrent work, SceneScape~\cite{fridman2023scenescape} and WonderJourney~\cite{yu2023wonderjourney} firstly generate 3D point cloud for scene by zoom-out and inpainting strategy. 2D image sequences are further rendered based on the reconstructed 3D point cloud. However, the accuracy of the 3D model critically impacts performance, particularly with novel camera trajectories.

% diffusion-based text to 3d generation.
% \vspace{-0.3cm}
\paragraph{Text-to-3D generation.}

% Recent text-to-3D methods \cite{lin2023magic3d, melas2023realfusion, metzer2023latent, wang2023prolificdreamer, poole2022dreamfusion} combine large text-to-image diffusion models \cite{rombach2022high} and neural radiance fields \cite{mildenhall2021nerf} to generate 3D objects without training. Several methods \cite{tang2023make, liu2023one, liu2023zero, rockwell2021pixelsynth} have been proposed that perform novel view synthesis from a single image. However, these methods typically either only work well on single objects, or restrict camera motions to small regions around the reference view. Recently, Text2room~\cite{hollein2023text2room} proposes a method to generate 3D indoor scenes based on text prompts. However, this work focuses specifically on creating indoor room mesh. In contrast, our \textit{DreamDrone} can generate endless indoor or outdoor scenes with various styles, such as Lego or van Goah.
Several text-to-3D generation methods~\cite{bautista2022gaudi, mo2023dit, chen2019text2shape, nichol2022point, zeng2022lion, kong2023priority} apply text-3D pair databases to learning a mapping function. However, supervised strategies remain challenging due to the lack of large-scale aligned text-3D pairs. CLIP-based~\cite{radford2021learning} 3D generation methods~\cite{zhang2022pointclip, mohammad2022clip, lee2022understanding, jiang20233d, jain2022zero} apply pre-trained CLIP model to create 3D objects by formulating the generation as an optimization problem in the image domain.
Recent text-to-3D methods like \cite{lin2023magic3d, melas2023realfusion, metzer2023latent, wang2023prolificdreamer, poole2022dreamfusion, shen2023anything, yang2024hash3d} blend text-to-image diffusion models \cite{rombach2022high} with neural radiance fields \cite{mildenhall2021nerf} for training-free 3D object generation. Other approaches \cite{tang2023make, liu2023one, liu2023zero, rockwell2021pixelsynth} focus on novel view synthesis from a single image, often limited to single objects or small camera motion ranges. Text2room~\cite{hollein2023text2room} generates 3D indoor scenes from text prompts, but is confined to room meshes.

% 与video generation的区别
% \vspace{-0.3cm}
\paragraph{Text-to-video generation.}

Generating videos from textual descriptions \cite{luo2023videofusion, hong2022cogvideo, ho2022video, blattmann2023align, singer2022make, zhang2023show, wang2023lavie, tan2024video} poses significant challenges, primarily due to the scarcity of high-quality, large-scale text-video datasets and the inherent complexity in modeling temporal consistency and coherence. CogVideo~\cite{hong2022cogvideo} addresses this by incorporating temporal attention modules into the pre-trained text-to-image model CogView2~\cite{ding2022cogview2}. The video diffusion model \cite{ho2022video} employs a space-time factorized U-Net, utilizing combined image and video data for training. Video LDM~\cite{blattmann2023align} adopts a latent diffusion approach for generating high-resolution videos. However, these methods typically do not account for the underlying 3D scene geometry in scene-related video generation, nor do they offer explicit control over camera movement. Additionally, their reliance on extensive training with large datasets can be prohibitively costly. While T2V-0~\cite{khachatryan2023text2video} introduced the concept of zero-shot text-to-video generation, its capability is limited to generating a small number of novel frames, with diminished quality in longer video sequences.

%% file: sec/3_method.tex
\section{Method}

% 在最开始的时候也说一下我们的setting是逐帧的生成。given x啥的。
We formulate the task of perpetual view generation as follows: given a starting image $I$, we generate the next view image $I'$ corresponding to an arbitrary camera pose $\left\{\mathcal{R},\mathcal{T} \right\}$, where the camera pose can be specified or via user's control.

\subsection{Preliminaries}
% pnp
We implement our method based on the recent state-of-the-art text-to-image diffusion model (\textit{i.e.} Stable Diffusion \cite{rombach2022high}). Stable diffusion is a latent diffusion model (LDM), which contains an autoencoder $\mathcal{D}(\mathcal{E}(\cdot )))$ and a U-Net \cite{ronneberger2015u} denoiser. Diffusion models are founded on two complementary random processes. The \textit{DDPM forward} process, in which Gaussian noise is progressively added to the latent code of a clean image: $\bm{x}_0$:
\begin{equation}
    \bm{x}_t = \sqrt{\alpha_t}\bm{x}_0 + \sqrt{1-\alpha_t}\bm{z},
\end{equation}
where $\bm{z}\sim \mathcal{N}(0, \mathbf{I})$ and $\left \{ \alpha_t \right \}$ are the noise schedule.

The \textit{backward} process is aimed at gradually denoising $\bm{x_T}$, where at each step a cleaner image is obtained. This process is achieved by a U-Net $\bm{\epsilon}_{\theta}$ that predicts the added noise $\bm{z}$. Each step of the backward process consists of applying $\bm{\epsilon}_{\theta}$ to the current $\bm{x}_t$, and adding a Gaussian noise perturbation to obtain a cleaner $\bm{x}_{t-1}$.

Classifier-guided DDIM sampling \cite{dhariwal2021diffusion} aims to generate images from noise conditioned on the class label. Given the diffusion model $\bm{\epsilon}_{\theta}$, the latent code $\bm{x}_t$ at timestep $t$, the classifier $p_{\theta}(y|x_t)$, and the gradient scale $s$, the sampling process for obtaining $\bm{x}_{t-1}$ is formulated as:
\begin{equation}
    \hat{\epsilon } = \bm{\epsilon}_{\theta}(\bm{x}_t) - \sqrt{\bar{\alpha}_{t-1}}\triangledown _{\bm{x}_t} \log p_{\phi}(y|\bm{x}_t),
    \label{eq:eps}
\end{equation}
and 
\begin{equation}
    \bm{x}_{t-1} = \sqrt{\bar{\alpha}_{t-1}} \cdot  \frac{\bm{x}_t - \sqrt{1-\bar{\alpha}_t}\hat{\epsilon }}{\sqrt{\bar{\alpha}_t}} + \sqrt{1-\bar{\alpha}_{t-1}}\hat{\epsilon },
    \label{eq:xt-1}
\end{equation}
where $\alpha$ is the denoise schedule.
% Diffusion models are rapidly evolving and have been extended and trained to progressively generate images conditioned on a guiding signal $\bm{\epsilon}_{\theta}(\bm{x}_t, \bm{y}, t)$, \textit{e.g.}, conditioning the generation on another image \cite{brooks2023instructpix2pix, tumanyan2023plug}, class label \cite{dhariwal2021diffusion}, or text \cite{kim2022diffusionclip, nichol2021glide, ramesh2022hierarchical, rombach2022high}.

% TokenFlow
In the self-attention block of the U-Net, features are projected into queries $\mathit{\mathbf{Q}}$, keys $\mathit{\mathbf{K}}$, and values $\mathit{\mathbf{V}}$. The output of the block $\bm{o}$ is obtained by: 
\begin{equation}
    \bm{o} = \mathbf{A}\mathbf{V}, \; \; \textup{where} \; \mathbf{A} = \textup{Softmax}(\mathit{\mathbf{Q}}\mathit{\mathbf{K}}^{\top })
    \label{eq:attn}
\end{equation}
The self-attention operation allows for long-range interactions between image tokens.

% \vspace{-0.3cm}
\subsection{Overview}
% \vspace{-0.6cm}

% 对于无限场景生成来说，最简单的步骤是1. 根据当前帧的RGBD和相机外参，获得下一帧的RGB。2. 利用inpainting和outpainting方法来补全下一帧图片存在的空洞。但是这样做会有如下问题：1. 需要填补的空洞很不规律，而现有的inpainting方法很难保证填补出来的区域符合整张图的语义。2. 由于深度并不是完全准确，导致warp后的图片某些区域会存在一些畸变。在随着相机移动的过程中，这些畸变会慢慢积累，最终导致生成的图片质量很差。3. 在图片向前移动的过程中，某些图片区域需要随之增加细节，但inpainting和outpainting方法并不能完成super resolution的功能。

\begin{figure*}[htbp] 
\centering 
\includegraphics[width=1.0\textwidth]{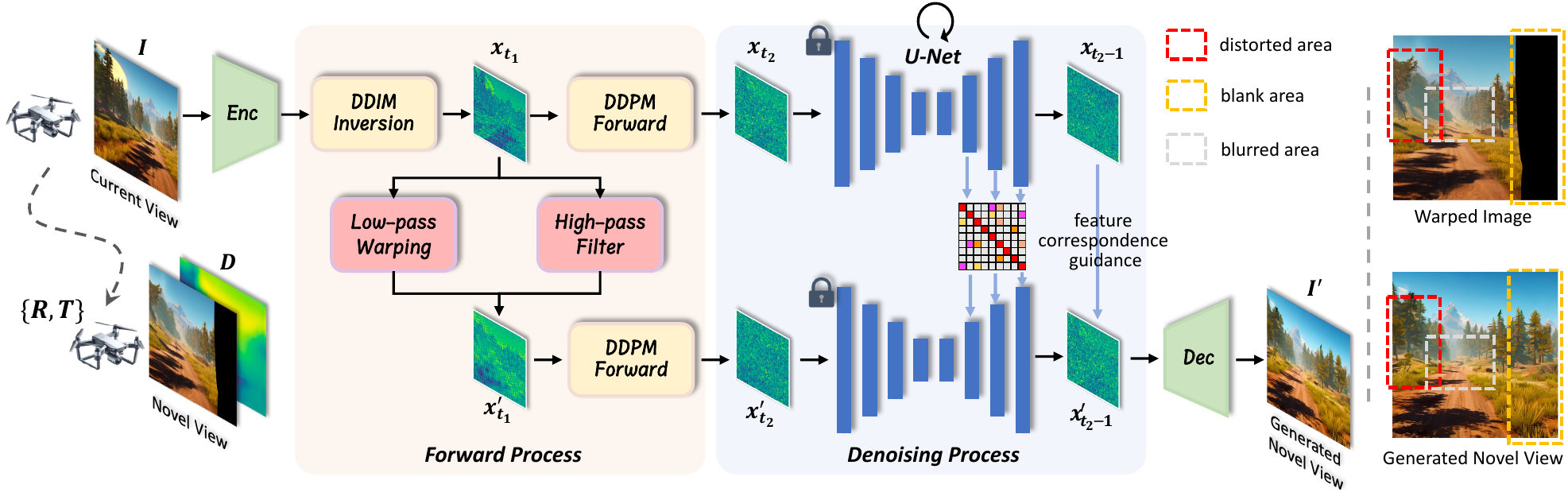} 
% \vspace{-0.3cm}
\caption{\textbf{Overview of our proposed pipeline.} Starting from a real or generated RGBD ($I$, $D$) image at the current view, we apply DDIM inversion to obtain intermediate latent code $x_{t_1}$ at timestep $t_1$ using a pre-trained U-Net model. A warping with the high-pass filter strategy is applied to generate latent code for the next novel view. A few more DDPM forward steps from timestep $t_1$ to $t_2$ are applied for enlarging the degree of freedom \textit{w.r.t.} the warped latent code. In the denoising process, we apply pre-trained U-Net to generate the novel view from $x_{t_2}'$. The cross-view self-attention module and feature-correspondence guidance are applied to maintain the geometry correspondence between $x_{t_2}$ and $x_{t_2}'$. The right side shows the warped image and our generated novel view $I'$. Our method greatly alleviates blurring, inconsistency, and distortion. The overall pipeline is zero-shot and training-free.
}
\label{fig2}
\end{figure*}

% \vspace{-0.6cm}
% 
 % is a very challenging problem: as the camera travels, we must fill in unseen missing regions in a harmonious manner, and must resolve new details as scene content approaches the camera, all the while maintaining photo-realism and diversity. To this end, previous methods~\cite{li2022infinitenature,liu2021infinite,cai2023diffdreamer} train a refiner to add details and synthesize new content in areas that require inpainting or outpainting. Though achieving promising results, the trained refiner can only fit for scenes that align with the training dataset. 
% To this end, we modify the powerful pre-trained text-to-image diffusion model as a refiner. 
Perpetual view generation as the camera moves presents a complex challenge. This process involves seamlessly filling in unseen regions caused by image warping, adding details to objects as they come closer, while ensuring the imagery remains realistic and diverse. Prior works~\cite{li2022infinitenature,liu2021infinite,cai2023diffdreamer} have focused on training a refiner to enhance details and create new content for areas requiring inpainting or outpainting. These efforts have shown promising outcomes, yet the effectiveness of the refiner is generally limited to scenarios that align with the training dataset.

% 加notation，加details的框

Since diffusion models can generate high-quality large-variety images from random latent code, a direct solution arises: can we modify the powerful pre-trained text-to-image diffusion model as a refiner? Empirically, DDIM inversion strategy~\cite{khachatryan2023text2video,Tumanyan_2023_CVPR} can obtain the intermediate latent code at each timestep and the image can be reconstructed by those latent codes. To this end, we attempt to explicitly warp the latent code of the current view and generate the novel view by the pre-trained text-to-image diffusion model. 

Our overall pipeline is illustrated in Fig.~\ref{fig2}. Initially, we obtain the latent code $x_{t_1}$ of the current view's RGB image $I$ at timestep $t_1$ through the DDIM inversion process. We then warp the current frame's latent code $x_{t_1}$ to the next view $x'_{t_1}$ using depth information and camera extrinsic parameters. However, directly denoising from $x'_{t_1}$ to the image also suffers from blurry, which results in the non-integer pixel coordinates and the interpolation operation. More noise is added from $x'_{t_1}$ at timestep $t_1$ to $x'_{t_2}$ at timestep $t_2$ by DDPM forward operation, for generating high-quality images. The side effect of DDPM is the geometry inconsistency between adjacent views. To this end, we propose a feature-correspondence-guidance denoising strategy to enforce geometry consistency. Moreover, a high-pass filtering strategy is proposed to maintain the consistency of high-frequency details between adjacent views. Please refer to \cref{fig:ablation} for the motivation of our proposed modules. Our overall pipeline requires only a pre-trained text-to-image diffusion model and a depth estimation model, eliminating the need for any additional training or fine-tuning.

\subsection{Warping latent codes}
\setlength\intextsep{0.1cm}
\begin{algorithm}
\caption{Warping latent code with high-pass filter}
\begin{algorithmic}[1]
\Require $\bm{x}_t$ \Comment{latent code  at timestep $t$ of current view $c$}
\State $F(\bm{x}_t) \gets FFT(\bm{x}_t)$ \Comment{Apply Fast Fourier Transform}
\State Split $F(\bm{x}_t)$ into $F_{low}$ and $F_{high}$ using threshold $\sigma$
\State $\bm{x}_t^{low} \gets IFFT(F_{low})$ \Comment{Inverse FFT on low-frequency component}
\State $\bm{x}_t^{low-warped} \gets warp(\bm{x}_t^{low})$ \Comment{warp the low-frequency content}
\State $F_{warped} \gets FFT(\bm{x}_t^{low-warped})$ \Comment{FFT on warped content}
\State $F' \gets F_{warped}+F_{high}$ \Comment{Combine low-frequency of warped content with high-frequency of original content}
\State $\bm{x}_t' \gets IFFT(F')$ \Comment{Inverse FFT to get latent code for next view $c'$}

\Return $\bm{x}_t'$ \Comment{warped latent code at timestep $t$ for next view $c'$} 
\end{algorithmic}
\label{alg:warp}
\end{algorithm}

The results in the right side of \cref{fig2} reveal that directly warping images based on camera intrinsics $K$, extrinsics $\left\{\mathcal{R},\mathcal{T} \right\}$, and depth information leads to regions of distortion in the images. Additionally, the use of inpainting \cite{rombach2022high, lugmayr2022repaint, yu2023inpaint} and outpainting \cite{li2023panogen, yang2022scene,yu2024shadow} models to fill these gaps does not achieve satisfactory outcomes. In pursuit of photo-realistic images, we opt to edit the latent code corresponding to timestep $t$. PnP \cite{tumanyan2023plug} and DIFT \cite{tang2023emergent} have shown that the features of diffusion possess strong semantic information, with semantic parts being shared across images at each step. The simplest method for warping the latent code follows the same approach as warping the image. The only difference between warping the latent code and warping the image is a slight modification in the camera intrinsics; this entails scaling the camera intrinsics proportionally based on the different resolutions of the image and latent code.

The overall procedure for warping the latent code is illustrated in Alg.~\ref{alg:warp}. Initially, a latent code \( {x}_t \) is obtained and transformed via Fast Fourier Transform (FFT) to \( F(\bm{x}_t) \). This is divided into low-frequency \( F_{\text{low}} \) and high-frequency \( F_{\text{high}} \) components, segregated at threshold \( \sigma \). The key step involves warping the Inverse FFT (IFFT) processed low-frequency component \( \bm{x}_t^{\text{low}} = \text{IFFT}(F_{\text{low}}) \), warping to the next view \( \bm{x}_t^{low-warped} \). Merging \( \text{FFT}(\bm{x}_t^{low-warped} )\) with \( F_{\text{high}} \), we obtain \( F' \), from which the final latent code \(\bm{x}_t' = \text{IFFT}(F') \) is reconstructed. This approach efficiently preserves high-frequency details, enabling high-fidelity scene generation aligned with text prompts.

% \vspace{-0.3cm}
\subsection{Feature-correspondence-guidance design}
% dragon diffusion
% 在获得下一帧对应的latent code$x_t'$之后，我们使用DDPM方法来增加latent code的自由度，以便生成更丰富的图片细节。增加自由度会带来一个问题，那便是帧与帧之间的相关性，无约束的diffusion去噪过程会使得相邻帧的语义相关性很差。基于此，我们提出了一种cross-view self attention策略，并通过计算相邻帧在feature层面的correspondence，进而引导denoising过程。下面我们会分别介绍。

After obtaining the latent code $x_t'$ corresponding to the next frame, we employ the DDPM (Denoising Diffusion Probabilistic Models) method to increase the degrees of freedom of the latent code, enabling the generation of richer image details. However, increasing freedom introduces a challenge: the correlation between frames. An unconstrained diffusion denoising process can result in poor semantic correlation between adjacent frames. To address this, we propose a feature-correspondence guidance strategy with a cross-view self-attention mechanism. We introduce these approaches in detail below.

% In this part, inspired by classifier guidance \cite{dhariwal2021diffusion}, we aim to obtain the intermediate latent code of the next camera view $\bm{x}_t'$ of the diffusion process by warping $\bm{x}_t$ at current camera view and denoising with , thereby achieving infinite and geometry-consistent scene generation.

% warp with fft

\paragraph{Cross-view self-attention.}
% injection, Dragon diffusion
To maintain consistency between the generated result and the original image, inspired by recent image and video editing works \cite{tumanyan2023plug,wu2023tune, geyer2023tokenflow, wang2023zero}, we modify the process of the self-attention module of U-Net when denoising the latent code $\bm{x}_t'$. Specifically, we denoise the views for the current and next view together. The key and value of the self-attention modules from the next view are replaced by that of the current view. To be specific, for obtaining the original view, the self-attention module is defined the same as \cref{eq:attn}. The modified cross-view self-attention for generating a novel view is defined as:
\begin{equation}
    \bm{o}' = \mathbf{A'}\mathbf{V}, \; \; \textup{where} \; \mathbf{A}' = \textup{Softmax}(\mathbf{Q}'\mathbf{K}^{\top }),
\end{equation}
where $\mathbf{Q}'$, $\mathbf{A}'$, and $\bm{o}'$ are query, attention matrix, and output features for the novel views. $\mathbf{K}$ and $\mathbf{V}$ are injected keys and values obtained from the self-attention module for generating the original view. Please note that the $\mathbf{K}$ and $\mathbf{V}$ are also warped before injection.

\paragraph{Feature-correspondence guidance.}
% Though the cross-view self-attention mechanism maintains geometry consistency between adjacent views, it is challenging to preserve high-frequency details when the camera is moving forward. The recent work, DIFT~\cite{tang2023emergent} demonstrates the strong correspondence of the intermediate features of the diffusion models, which can be used for point-to-point matching between different images. Moreover, vanilla classifier guidance \cite{dhariwal2021diffusion} applies gradients of the pre-trained classifier to guide the diffusion sampling process toward an arbitrary class label. To enforce the consistency between adjacent views, we introduce a feature correspondence guidance to the DDIM process.

Maintaining geometry consistency between adjacent views using the cross-view self-attention mechanism presents challenges, especially in preserving high-frequency details as the camera moves forward. The recent DIFT~\cite{tang2023emergent} highlights the potential of using intermediate features of diffusion models for accurate point-to-point image matching~\cite{qiu2019matching}. Additionally, the concept of vanilla classifier guidance \cite{dhariwal2021diffusion} steers the diffusion sampling process using pre-trained classifier gradients towards specific class labels. Building on these ideas, we integrate feature correspondence guidance into the DDIM sampling process to enhance consistency between adjacent views, addressing the challenge of detail preservation in dynamic scenes.

Specifically, we obtain the features of the current and next view at each timestep $t$ of the DDIM process and calculate the cosine distance between the warped original features and features from the next novel views:
\begin{equation}
    \mathcal{L}_{sim}^{t} = \frac{1 - \cos \left [ \mathrm{warp}(f_t), f_t'\right ]}{2},
\end{equation}
where $f_t$ and $f_t'$ are intermediate features extracted from pre-trained U-Net $\bm{\epsilon}_{\theta}$ at timestep $t$ and $\mathrm{warp}$ is the warping functions. The lower $\mathcal{L}_{sim}^{t}$, the higher the similarity.

We further introduce the similarity score $\mathcal{L}_{sim}^t$ to the DDIM sampling process, for generating novel views with geometry consistency. The predicted noise $\hat{\epsilon }$ is formulated as: 
\begin{equation}
    \hat{\epsilon } = \bm{\epsilon}_{\theta}(\bm{x}_t) - \lambda \sqrt{\bar{\alpha}_{t-1}}\triangledown _{\bm{x}_t} \mathcal{L}_{sim}^{t},
    \label{eq:eps}
\end{equation}
where $\lambda$ is the constant hyper-parameter and latent code $\bm{x}_{t-1}$ is calculated by \cref{eq:xt-1}

%% file: sec/4_experiments.tex
% \vspace{-0.3cm}
\section{Experiments}
\label{exp}

% \vspace{-0.2cm}
\subsection{Implementation details}
% \vspace{-0.3cm}
% text-to-video zero
We take Stable Diffusion \cite{rombach2022high} with the pre-trained weights from version 2.1\footnote{\href{https://huggingface.co/stabilityai/stable-diffusion-2-1-base}{https://huggingface.co/stabilityai/stable-diffusion-2-1-base}} as the basic text-to-image diffusion and MiDas~\cite{ranftl2020towards} with weights $\mathrm{dpt\_beit\_large\_512}$\footnote{\href{https://github.com/isl-org/MiDaS}{https://github.com/isl-org/MiDaS}}. The overall diffusion timesteps is $1000$. We warp the latent code at timestep $t_1$=$21$ and add more degrees of noise to timestep $t_2$=$441$. The threshold $\sigma$ for the high-pass filter is $20$ and the hyper-parameter $\lambda$ for feature-correspondence guidance is $300$.
Due to the page limit, please refer to the supplementary material (\textit{supp.}) for details.

% \vspace{-0.5cm}
\subsection{Baselines} 
% \vspace{-0.3cm}
We compare against 1) two supervised methods for perpetual view generation: \textit{InfNat} \cite{liu2021infinite} and \textit{InfNat-0} \cite{li2022infinitenature}. 2) one text-conditioned 3D point cloud-based scene generation: \textit{SceneScape}~\cite{fridman2023scenescape}. 3) two supervised methods for text-to-video generation: \textit{CogVideo} \cite{hong2022cogvideo} and \textit{VideoFusion} \cite{luo2023videofusion}. 4) one method for zero-shot text-to-video generation: \textit{T2V-0} \cite{khachatryan2023text2video}.

% \begin{itemize}
%     \item \textit{InfNat} \cite{liu2021infinite}: 
%     \item \textit{InfNat-0} \cite{li2022infinitenature}: 
%     \item \textit{CogVideo} \cite{hong2022cogvideo}: 
%     \item \textit{VideoFusion} \cite{luo2023videofusion}:
%     \item \textit{Text2Video-Zero} \cite{khachatryan2023text2video}: 
% \end{itemize}

% text2room, text2video zero, make-a-video, tokenflow
% \vspace{-0.3cm}
% \vspace{-0.5cm}
\subsection{Evaluation metrics} 
% \vspace{-0.3cm}
We evaluate our zero-shot perpetual scene generation into two aspects: 1) the quality of generated images and text-image alignment, and 2) the temporal consistency of generated image sequences.
% \vspace{-0.3cm}
\paragraph{Image quality and text-image alignment.} We evaluate CLIP score~\cite{radford2021learning}, which indicates text-scene alignment for quantitative comparisons. 
A high average CLIP score indicates not only that the generated images are more aligned with the corresponding prompts but also that they consistently maintain high quality~\cite{khachatryan2023text2video}. 
CogVideo~\cite{hong2022cogvideo}, VideoFusion~\cite{luo2023videofusion}, SceneScape~\cite{fridman2023scenescape}, and T2V-0~\cite{khachatryan2023text2video} are all engaged in text-conditioned generation tasks. We generated 50 scene-related text prompts using GPT-4\footnote{\href{https://openai.com/gpt-4}{https://openai.com/gpt-4}} and then created videos using each of the three methods. For the InfNat~\cite{liu2021infinite} and InfNat-0~\cite{li2022infinitenature} methods, we used Stable Diffusion to generate the initial frame, followed by subsequent frame generation based on this initial frame. We calculated the distance between each generated frame and the text embedding, known as the CLIP score. Considering that the InfNat~\cite{liu2021infinite} and InfNat-0~\cite{li2022infinitenature} methods trained on natural scene datasets, we further provided 10 very general prompts such as `an image of the landscape' and `an image of the mountain' for these methods, and then selected the highest CLIP score as the CLIP score for the current frame.
% clip score -- 10 frames, 20 frames, 100 frames.
% \vspace{-0.3cm}

\begin{table}[htbp]
  \centering
  \captionsetup{font=footnotesize}
  \caption{\textbf{Ablations of image quality and temporal coherence of generated image sequences with various lengths.} Please refer to \cref{fig:ablation} for quality comparisons.}
  % \vspace{-0.3cm}
  \scalebox{0.7}{
  
    \begin{tabular}{c|ccc|ccc|ccc}
    \hline
    \multirow{2}[1]{*}{Methods} & \multicolumn{3}{c|}{PSNR $\uparrow$} & \multicolumn{3}{c|}{SSIM $\uparrow$} & \multicolumn{3}{c}{CLIP $\uparrow$} \\
          & 8 frames & 16 frames & 32 frames & 8 frames & 16 frames & 32 frames & 8 frames & 16 frames & 32 frames \\
      \hline
    \rowcolor{lightgray!30}warp image &26.90 & 22.46 & 21.62 & 0.25  & 0.23  & 0.24  & 0.138 & 0.112 & 0.106 \\
    warp latent &28.35 & 28.57 & 28.75 & 0.27  & 0.28  & 0.24  & 0.135 & 0.122 & 0.125 \\
    \rowcolor{lightgray!30}warp latent+DDPM &24.67 & 23.04 & 22.59 & 0.12  & 0.10  & 0.06  & 0.302 & 0.297 & 0.308 \\
    warp latent+DDPM+guidance & 28.27 & 28.21 & 28.10 & 0.34  & 0.30  & 0.26  & 0.317 & 0.316 & 0.313 \\
    \rowcolor{lightgray!30}\makecell[c]{warp latent+DDPM+guidance\\+cross-view attn.} & 28.89 & 28.83 & 28.75 & 0.32  & 0.31  & 0.27  & 0.318 & 0.315 & 0.315 \\
    % \hline
    \makecell[c]{warp latent+DDPM+guidance\\+cross-view attn.+high pass filter}  &   29.91    &    29.86   &   29.79    &    0.39   &    0.38   &     0.35  &   0.320    &   0.318    &  0.319\\
    \hline
    \end{tabular}
    }
    
  \label{tab:ablation}%
\end{table}%

\paragraph{Temporal consistency of generated image sequences.} We demonstrate our advancements in temporal consistency against other SOTA methods by calculating average PSNR and SSIM scores across adjacent frames for generated videos with different lengths. The higher scores demonstrate the superiority in terms of cross-view consistency.

% \vspace{-0.4cm}
\subsection{Ablation studies}
\begin{figure}[htbp] 
\centering 
\includegraphics[width=0.85\textwidth]{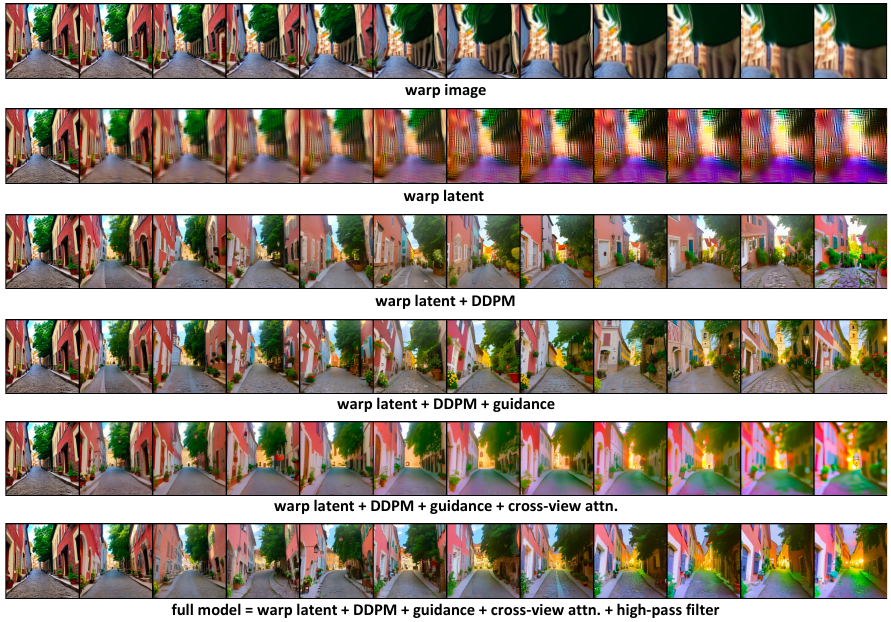} 
% \vspace{-0.3cm}
\captionsetup{font=footnotesize}
\caption{\textbf{Ablation results for the key components.} We perform ablation studies by disabling the key components of our method. We illustrate every five frames for each ablation experiment. Please zoom in for better comparisons.}
\label{fig:ablation}
\end{figure}

% \vspace{-0.2cm}
We perform ablation studies on our three proposed modules: 1) warping latent with high-pass filter, 2) cross-view self-attention module, and 3) feature-correspondence guidance. The quantitative ablation results are shown in \cref{tab:ablation} and we visualize the ablation samples in \cref{fig:ablation}.

% 无线场景生成任务最简单的解决方案就是直接逐帧的warp image。但这种方法显然是不可行的。由于在warp image的过程中会引入畸变等误差，且随着相机的推进，图片里很多细节消失，这会导致warp的图片越来越模糊。从tab 1第一行可以看出，其CLIP score非常低，代表其图片质量很差。Fig 4第一行结果也可佐证这一点。接下来我们探究了DDPM在整个pipeline是否必要。我们引入ddpm的目的是通过增加噪声来增加diffusion模型的自由度。Warping imgs or latent codes results in non-integer pixel coordinates, leading to interpolation-induced blurring. 所以如果不加入DDPM，会使得生成的图片越来越模糊。Please refer to 2nd row in Fig 4.

% 无限场景生成任务最简单的方法就是逐帧的warp image，但是这种方法是不可行的。Warping images results in non-integer pixel coordinates, leading to interpolation-induced blurring and distortion. 且这种误差会随着逐帧的生成而逐渐累积，进而导致质量崩塌。Tab 1的前两行表明直接warp image或warp latent code (即去掉DDPM）的CLIP score非常低，说明生成图片的质量很差。从图4的前两行也可以看到生成的图片越来越模糊。我们通过引入DDPM来增加diffusion模型的自由度，从而生成高质量的图片。但引入DDPM带来的副作用是使得相邻帧的语义一致性变差。please refer to the supp for the generated results with different scale of DDPM forward process.

The simplest method for infinite scene generation tasks is frame-by-frame image warping, but this approach is unfeasible, as is directly warping the latent code. Warping images leads to non-integer pixel coordinates, resulting in interpolation-induced blurring and distortion. Moreover, these errors accumulate with each frame generated, leading to a collapse in quality. The first two rows of Table 1 show that directly warping images or warping latent codes (i.e., removing DDPM) results in very low CLIP scores, indicating poor quality of the generated images. The generated images becoming progressively blurred can also be observed in the first two rows of \cref{fig:ablation}. We introduce DDPM to increase the degrees of freedom of the diffusion model, thereby generating high-quality images. However, the introduction of DDPM has the side effect of worsening the semantic consistency between adjacent frames ($3^{rd}$ row in \cref{tab:ablation} and \cref{fig:ablation}).With the help of DDPM, the CLIP score increases from $0.125$ to $0.308$ when generating 32 images. Please refer to the \textit{supp.} for the generated results with different scales of the DDPM forward process.

% 

% Without feature-correspondence guidance and cross-view attention, even if we perform the corresponding warpping at the feature level, it is challenging to ensure strong consistency between generated adjacent frames ($1^{st}$ row). Direct warpping of the latent code leads to increasingly blurred images and a loss of high-frequency details. However, the use of our low-pass warpping strategy effectively ensures that the generated images retain high-definition details ($2^{nd}$ row). Without the guidance of the feature-correspondence guidance mechanism or the presence of a cross-view self-attention module, it is difficult to guarantee the geometric consistency of adjacent frames ($3^{rd}$ and $4^{th}$ rows). The components in our proposed pipeline enhance each other, ensuring that each generated frame is not only rich in detail but also maintains good consistency with adjacent frames. Please refer to \textit{supp.} for more detailed ablation studies.

% 为了在保证图片生成质量的同时兼顾consistency with adjacent views，我们提出了feature-correspondence guidance策略。对比图4的第三行和第四行可以看出，在增加guidance后，相邻帧的语义一致性明显增强, PSNR和SSIM分数也都有着明显的提升。为了进一步提升cross-view consistency，我们采用了cross-view attention modules和low-pass filtering。从可视化结果可以看出，在增加了cross-view attention module后，相邻相机视角的语义一致性进一步增强。low-pass filter操作进一步保留了当前帧的高频细节，从而进一步提升了相邻帧高频细节的语义一致性。

To ensure the quality of image generation while also maintaining consistency with adjacent views, we propose a feature-correspondence guidance strategy. Comparing the third and fourth rows of \cref{fig:ablation}, it is evident that the semantic consistency between adjacent frames is significantly enhanced after adding guidance, with noticeable improvements in both PSNR and SSIM scores in \cref{tab:ablation}. To further enhance cross-view consistency, we adopted cross-view attention modules and high-pass filtering. From the visualized results at the $5^{th}$ and $6^{th}$ rows in \cref{fig:ablation}, it is clear that the semantic consistency of adjacent camera perspectives is further strengthened after incorporating the cross-view attention module. The operation of the high-pass filter further preserves the high-frequency details of the current frame, thereby further enhancing the semantic consistency of high-frequency details between adjacent frames. For instance, comparing the left side house at the $4^{th}$, $5^{th}$, and $6^{th}$ rows in \cref{fig:ablation}, the cross-view consistency is enhanced after adding the proposed modules.

% 没有feature-correspondence guidance和cross-view attention时，即使我们在feature层面做了对应的warpping，但很难保证生成的相邻帧之间保持很强的一致性（第一行）。如果直接对latent code做warpping，会导致生成的图越来越模糊，越来越缺少高频细节。而在使用low-pass warpping策略后，能够很好的保证生成图有着高清的细节（第二行）。在没有feature-correspondence guidance机制的引导下或没有cross-view self-attention模组的情况下，很难保证相邻帧的几何一致性（第三四行）。我们提出的pipeline中的模组之间彼此促进，促使生成的每一帧图片既有着丰富的细节，相邻帧之间也有着很好的一致性。

% 除了对我们提出的modules进行消融实验外，我们继续尝试回答几个问题：Q1. 在warping latent Code之后，引入DDPM forward操作是否必要？Q2. 显式的warp latent code是否可以控制相机视角的运动轨迹？ Q3. 我们的方法可以通过改变text prompts从一个场景穿梭到另一个场景么？

In addition to conducting ablation experiments on the modules we propose, we continue to explore two more questions:

\begin{figure}[htbp] 
\centering 
\captionsetup{font=footnotesize}
\includegraphics[width=0.99\textwidth]{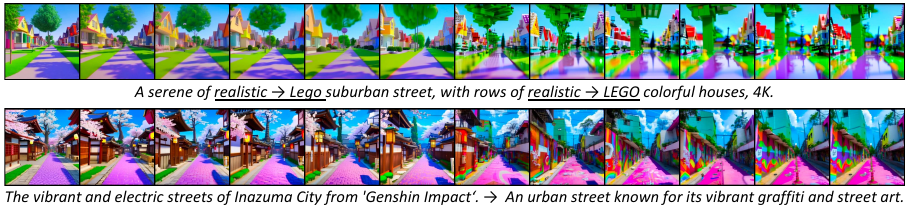} 
% \vspace{-0.3cm}
\caption{\textbf{Ablation study for scene travel.} We visualize two image sequences and change the prompt when generating novel views. We illustrate every five images and the prompts are changed when generating $31^{th}$ image ($7^{th}$ image shown in each row).}
\label{fig:change_prompt}
\end{figure}

% 时序不一致性是有什么引起的？

% \paragraph{Q1:} \textit{Is it necessary to introduce DDPM forward operations after warping the latent code?}

% 我们在逐帧的过程中改变了textual prompts，生成结果如图5所示。从可视化结果可以看出，DreamDrone可以在保证相邻帧的语义一致性的情况下，根据prompt的变化，非常平滑的完成了场景的转换(streets in Inazuma City -> urban art street)或场景风格的转换（realistic -> lego style)。
% \vspace{-0.3cm}

\textit{Q1: Can DreamDrone shuttle from one scene to another by changing text prompts?} During the frame-by-frame process, we changed the textual prompts, with the generated results shown in \cref{fig:change_prompt}. The visualized results demonstrate that DreamDrone can smoothly complete the scene travel (from streets in Inazuma City to urban art street) or the transition of scene styles (from realistic to Lego style) while ensuring the semantic consistency of adjacent views, according to the changes in textual prompts.

\begin{figure}[htbp] 
\centering 
\includegraphics[width=0.99\textwidth]{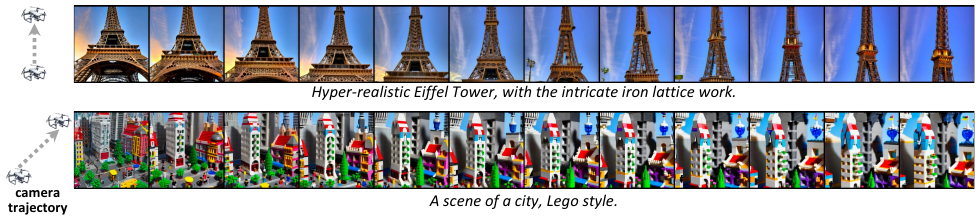} 
% \vspace{-0.3cm}
\captionsetup{font=footnotesize}
\caption{\textbf{Ablation study on customized camera trajectory.} We generate images with different camera directions. For the sample of the Eiffel Tower, our camera perspective continuously ascends. For the $2^{nd}$ scene of the Lego city, our camera not only moves forward but also shifts upwards and to the right.}
\label{fig:camera_trajectory}
\end{figure}

\textit{Q2: Can explicitly warping the latent code control the trajectory of camera perspective movement? } Since our method generates image sequences frame by frame, we can freely adjust the camera's flight angle by altering the camera's extrinsic parameters. In \cref{fig:camera_trajectory}, we provide sequences of images generated under different camera trajectories. The results show that our method possesses a high degree of freedom, allowing for the free customization of the camera's trajectory. Other state-of-the-art methods cannot achieve this functionality.
\subsection{Qualitative comparison}
% \vspace{-0.1cm}

\begin{figure*}[htbp] 
\centering 
\includegraphics[width=0.9\textwidth]{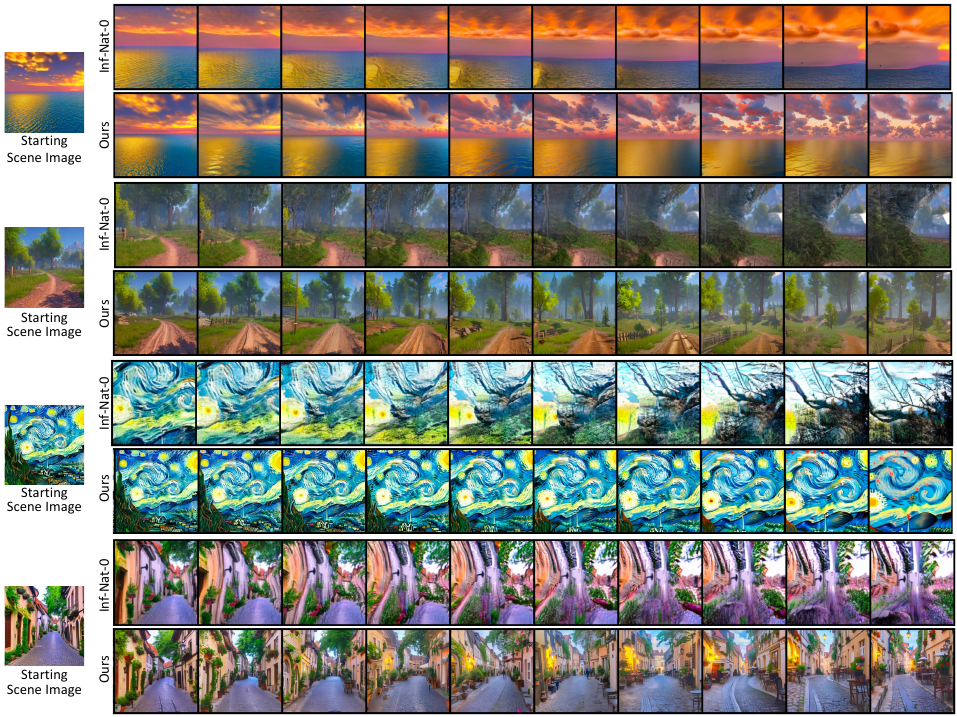} 
% \vspace{-0.3cm}
\captionsetup{font=footnotesize}
\caption{\textbf{Qualitative comparisons of InfNat-0~\cite{li2022infinitenature} and ours.} We provide four starting scene images with various styles and categories as start points and ask models to fly through the images. 50 frames are generated and we illustrate every five frames for each starting scene image.}
\label{inf}
\end{figure*}

In our comparison with InfNat-0~\cite{li2022infinitenature} (\cref{inf}), focusing on various scenes including coastlines, rivers, Van Gogh-style landscapes, and city streetscapes, we identified four main differences: Firstly, InfNat-0 shows proficiency in coastline scenes, a reflection of its training data, but our training-free \textit{DreamDrone} surpasses it in later frames due to InfNat-0's cumulative errors over time. Secondly, in natural scenes with closer objects, InfNat-0's flawed generation becomes more apparent, whereas our method maintains consistency. Thirdly, InfNat-0's limited approach to gap filling leads to poor performance in stylized scenes, in contrast to \textit{DreamDrone} which preserves high-frequency details and frame correspondence. Finally, in urban environments, InfNat-0 struggles significantly, while \textit{DreamDrone} achieves realistic and geometry-consistent views, demonstrating its versatility across varied scenarios.

\begin{figure*}[htbp] 
\centering 
\includegraphics[width=0.8\textwidth]{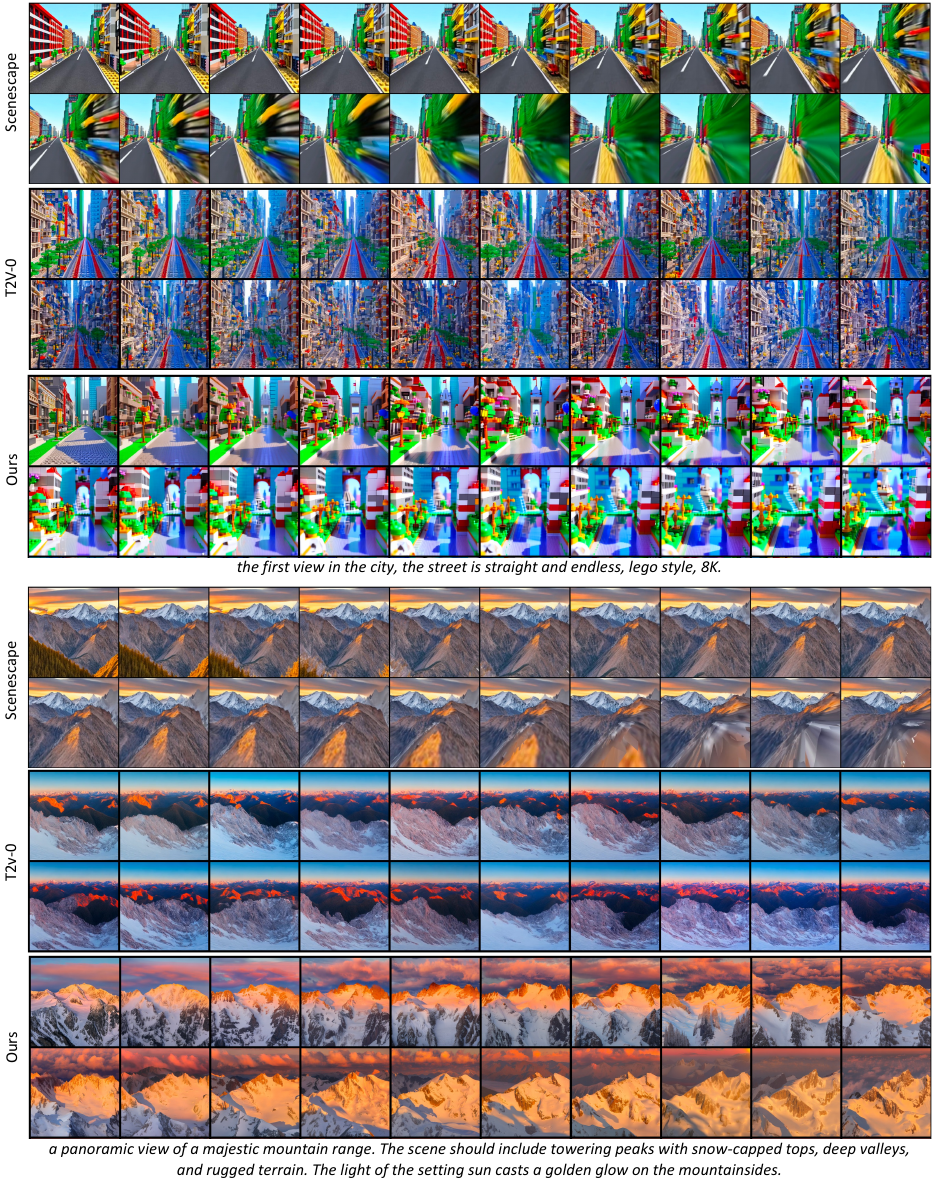} 
\captionsetup{font=footnotesize}
\caption{\textbf{Qualitative comparisons of SceneScape~\cite{fridman2023scenescape}, T2V-0~\cite{khachatryan2023text2video}, and our DreamDrone.} We visualize 20 continuous frames for each textual prompt. As the camera flies, our method generates geometry-consistent scene sequences.}
\label{t2v-0}
\end{figure*}

T2V-0~\cite{khachatryan2023text2video} introduces unsupervised text-conditioned video generation using stable diffusion. SceneScape~\cite{fridman2023scenescape} focuses on `zoom out' effects during backward camera movement. However, as seen in \cref{t2v-0}, both methods have limitations. SceneScape struggles with outdoor scenes and forward camera movement, leading to blurred and distorted results after 8 steps due to its reliance on a pre-trained inpainting model. T2V-0 displays a drop in quality beyond the third frame in complex environments like Lego-style cities, likely from its latent code editing approach that compromises frame continuity and geometric consistency. Conversely, our \textit{DreamDrone} excels across various scenes. It maintains detail, continuity, and quality in advancing camera scenarios, evident in even simpler landscapes like mountains where T2V-0 and SceneScape cannot effectively portray dynamic elements like cloud movement. Our approach ensures the preservation of fine details such as shadows and sunlight, creating a more dynamic and realistic video experience. Please refer to \textit{supp.} for more comparisons.

As our task bears similarities to text-to-video generation, we further provide qualitative comparisons with VideoFusion~\cite{luo2023videofusion}. Due to the page limit, please refer to \textit{supp.} for detailed comparisons. 

% \vspace{-0.3cm}
\subsection{Quantitative comparison}
% \vspace{-0.1cm}
\cref{tab:results} offers a detailed comparison of various SOTA methods for generating image sequences, including our method, DreamDrone. When compared to other training-based methods, DreamDrone, despite being training-free, consistently achieves higher CLIP scores across all frame lengths (0.320, 0.318, 0.319 for 8, 16, and 32 frames respectively). This is particularly noteworthy as the CLIP scores for training-based methods generally degrade as the number of generated frames increases. For instance, VideoFusion’s~\cite{luo2023videofusion} CLIP scores decrease from 0.281 for 8 frames to 0.272 for 32 frames. This trend suggests a decline in the quality of generated images with an increase in sequence length for training-based methods.

\begin{table}[htbp]
  \centering
  \captionsetup{font=footnotesize}
  \caption{\textbf{Qualitative comparisons with other SOTA methods.} We evaluate the quality and temporal coherence of the generated image sequences with various lengths.}
  % \vspace{-0.3cm}
    \scalebox{0.65}{
    \begin{tabular}{c|c|ccc|ccc|ccc}
    \hline
    \multirow{2}[1]{*}{} & \multirow{2}[1]{*}{Methods} & \multicolumn{3}{c|}{PSNR $\uparrow$} & \multicolumn{3}{c|}{SSIM $\uparrow$} & \multicolumn{3}{c}{CLIP $\uparrow$} \\
          &       & 8 frames & 16 frames & 32 frames & 8 frames & 16 frames & 32 frames & 8 frames & 16 frames & 32 frames \\
    \hline
    \cellcolor{white}\multirow{4}[1]{*}{training-based} & \cellcolor{gray!10}InfNat \cite{liu2021infinite} & \cellcolor{gray!10}28.75 & \cellcolor{gray!10}28.67 & \cellcolor{gray!10}28.65 & \cellcolor{gray!10}0.32  & \cellcolor{gray!10}0.30  & \cellcolor{gray!10}0.30  & \cellcolor{gray!10}0.125 & \cellcolor{gray!10}0.123 & \cellcolor{gray!10}0.118 \\
          & \cellcolor{white}InfNat-0 \cite{li2022infinitenature} & 28.92 & 28.89 & 28.87 & 0.37  & 0.35  & 0.34  & 0.128 & 0.125 & 0.122 \\
          & \cellcolor{gray!10}CogVideo \cite{hong2022cogvideo} & \cellcolor{gray!10}31.03 & \cellcolor{gray!10}30.08 & \cellcolor{gray!10}29.32 & \cellcolor{gray!10}0.45  & \cellcolor{gray!10}0.39  & \cellcolor{gray!10}0.31  & \cellcolor{gray!10}0.255 & \cellcolor{gray!10}0.249 & \cellcolor{gray!10}0.241 \\
          & \cellcolor{white}VideoFusion \cite{luo2023videofusion} & 29.89 & 28.36 & 28.78 & 0.41  & 0.37  & 0.31  & 0.281 & 0.283 & 0.272 \\
    \hline
    \cellcolor{white}\multirow{3}[4]{*}{training-free} & \cellcolor{gray!10}T2V-0 \cite{khachatryan2023text2video} & \cellcolor{gray!10}27.25 & \cellcolor{gray!10}26.17 & \cellcolor{gray!10}26.03 & \cellcolor{gray!10}0.27  & \cellcolor{gray!10}0.24  & \cellcolor{gray!10}0.23  & \cellcolor{gray!10}0.312 & \cellcolor{gray!10}0.305 & \cellcolor{gray!10}0.287 \\
          & \cellcolor{white}Scenescape \cite{fridman2023scenescape} & 29.87 & 29.75 & 29.66 & 0.41  & 0.38  & 0.34  & 0.318 & 0.282 & 0.279 \\
          & \cellcolor{gray!10}\textbf{DreamDrone (Ours)}  & \cellcolor{gray!10}29.91 & \cellcolor{gray!10}29.86 & \cellcolor{gray!10}29.79 & \cellcolor{gray!10}0.39  & \cellcolor{gray!10}0.38  & \cellcolor{gray!10}0.35  & \cellcolor{gray!10}0.320 & \cellcolor{gray!10}0.318 & \cellcolor{gray!10}0.319 \\
    \hline
    \end{tabular}%
    }
  \label{tab:results}%
\end{table}%

In contrast, DreamDrone maintains high CLIP scores even as the sequence length increases, indicating superior image quality. When compared to other training-free methods, DreamDrone also stands out. For example, while T2V-0’s~\cite{khachatryan2023text2video} CLIP scores decrease from 0.312 for 8 frames to 0.287 for 32 frames, DreamDrone’s CLIP scores remain relatively stable, further demonstrating its robustness in maintaining image quality across varying sequence lengths. This analysis underscores the effectiveness of DreamDrone in generating high-quality, temporally coherent image sequences without the need for training.

%% file: sec/5_conclusion.tex
\section{Conclusion}

In this work, we propose \textit{DreamDrone}, a novel approach for generating flythrough scenes from textual prompts without the need for training or fine-tuning. Our method explicitly warps the intermediate latent code of a pre-trained text-to-image diffusion model, enhancing the quality of the generated images and the generalization ability. We propose a feature-correspondence-guidance diffusion process and a high-pass filtering strategy to ensure geometric and high-frequency detail consistency. Experimental results indicate that \textit{DreamDrone} surpasses current methods in terms of visual quality and authenticity of the generated scenes.

\section*{Acknowledgement}
This project is supported by the Ministry of Education, Singapore, under its Academic Research Fund Tier 2 (Award Number: MOE-T2EP20122-0006).

%% file: sec/X_suppl.tex
% \clearpage
% \setcounter{page}{1}
% \maketitlesupplementary
% \begin{abstract}
    
% % 在该补充材料中，我们提供了更加全面的ablation studies，与text-to-video方法的可视化结果对比和我们方法更多的可视化结果，最后讨论了该方法的局限性。
% In this supplementary material, we provide more comprehensive ablation studies, comparisons of visual results with text-to-video methods, and additional visual results of our method. Finally, we discuss the limitations of this approach.

% \end{abstract}

In this supplementary material, we provide more comprehensive ablation studies, comparisons of visual results with text-to-video methods, and additional visual results of our method. Finally, we discuss the limitations and social impact of this approach.

\section{Implementation details}
\label{supp:implementation}

We take Stable Diffusion \cite{rombach2022high} with the pre-trained weights from version 2.1\footnote{\href{https://huggingface.co/stabilityai/stable-diffusion-2-1-base}{https://huggingface.co/stabilityai/stable-diffusion-2-1-base}} as the basic text-to-image diffusion and MiDas~\cite{ranftl2020towards} with weights $\mathrm{dpt\_beit\_large\_512}$\footnote{\href{https://github.com/isl-org/MiDaS}{https://github.com/isl-org/MiDaS}}. The overall diffusion timesteps is $1000$. We warp the latent code at timestep $t_1$=$21$ and add more degrees of noise to timestep $t_2$=$441$. The threshold $\sigma$ for high-pass filter is $20$ and the hyper-parameter $\lambda$ for feature-correspondence guidance is $300$. We conducted the experiments on Titan-RTX GPU. The generated speed is roughly 15 seconds per image.

\section{Ablation studies}
\label{supp:ablation}
% 首先，我们来探究在我们的pipeline里warp latent code的timestep t1对生成效果的影响，对比结果请见图1。根据diffusion的原理，较小的timestep意味着其对应的latent code的noise不是很多，故该latent code具有着较为完整的content和detail信息。当我们在较小的timestep（例如t1=1或21时）warp latent code，可以看到随着相机的移动，DreamDrone还是能够很好的保持其较为完整的content信息。随着timestep的增大，噪声较多，content的信息被一定程度的破坏掉，这时我们再warp latent，就不能得到较好的生成结果了。该实验表明，较小的timestep对应的latent code有着更为丰富的geometry和content 信息。
\begin{figure}[htbp] 
\centering 
\includegraphics[width=0.99\textwidth]{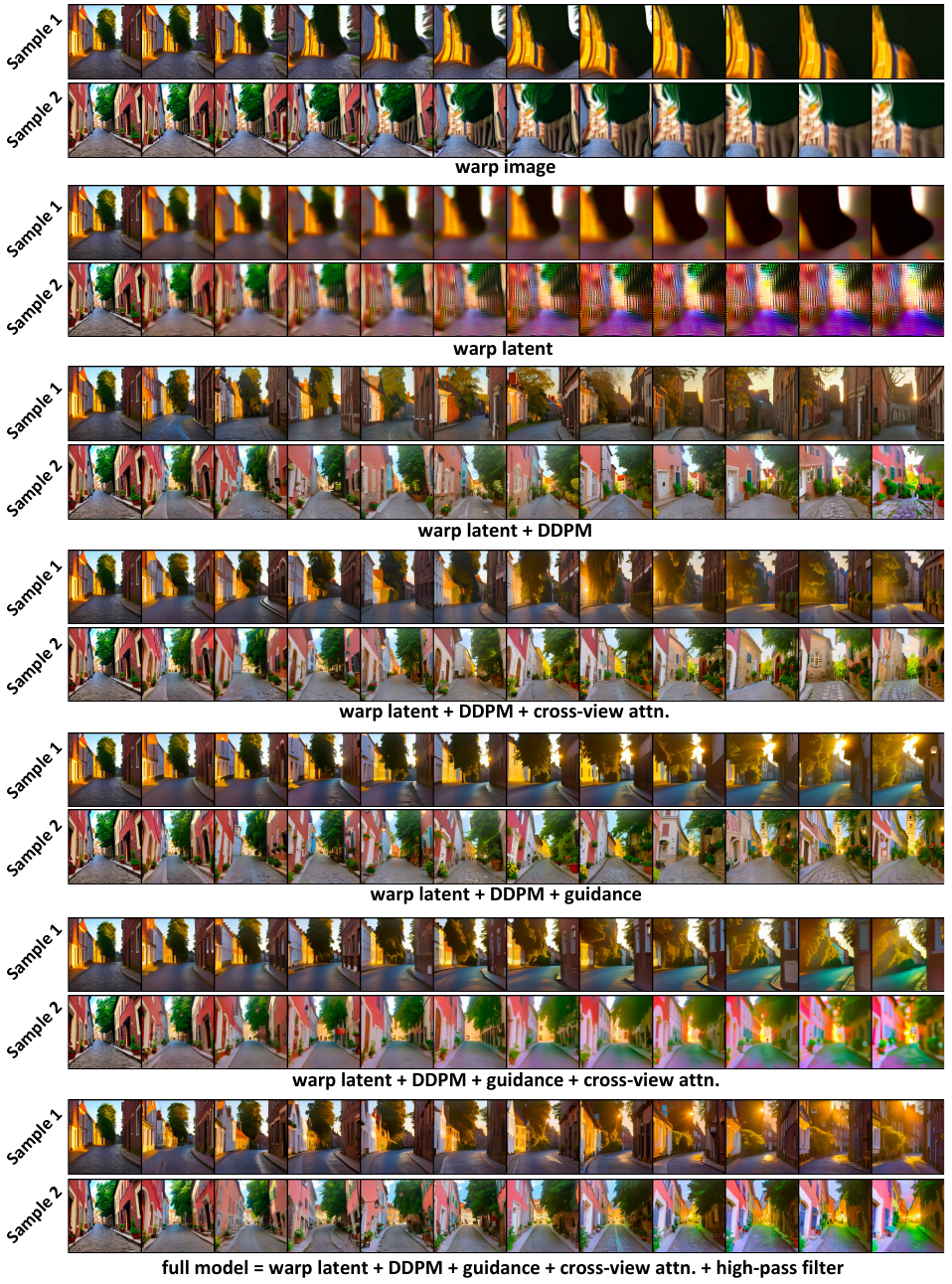} 
\vspace{-0.3cm}
\captionsetup{font=footnotesize}
\caption{\textbf{Ablation results for the key components.} We perform ablation studies by disabling the key components of our method. We illustrate every five frames for each ablation experiment. Please zoom in for better comparisons.}
\label{fig:ablation-full}
\end{figure}

In this section, we first provide additional ablation results for \cref{fig:ablation-full} as referenced from Fig.~3 in the main text. To enhance the robustness of our ablations, we include one example for each experimental setup. The comprehensive results of the ablation study for each component are illustrated in \cref{fig:ablation-full}. For quantitative results, please refer to Tab.~1 in the main text. The simplest method for tasks involving infinite scene generation is frame-by-frame image warping. However, this approach is impractical, as is the direct warping of the latent code. Warping images results in non-integer pixel coordinates, which leads to interpolation-induced blurring and distortion. Furthermore, these errors accumulate with each generated frame, causing a significant degradation in quality, with the images becoming progressively blurred, as shown in \texttt{warp image} and \texttt{warp latent} in \cref{fig:ablation-full}.

To address these challenges, we introduce DDPM to increase the degrees of freedom for the diffusion model, facilitating the generation of high-quality images (\texttt{warp latent + DDPM}). With the incorporation of DDPM, the CLIP score improves from $0.125$ to $0.308$ for a series of 32 images. Nevertheless, the introduction of DDPM inadvertently affects the semantic consistency between adjacent frames. 

To maintain the integrity of image content, and inspired by previous methods~\cite{tumanyan2023plug,khachatryan2023text2video}, we integrate cross-view attention modules into our framework. As demonstrated in \texttt{warp latent + DDPM + cross attn.} in \cref{fig:ablation-full}, the consistency of geometries across views is significantly improved compared to \texttt{warp latent + DDPM}. To ensure high-quality image generation while also maintaining consistency across adjacent views, we propose a feature-correspondence guidance strategy. Comparing \texttt{warp latent + DDPM + guidance} with \texttt{warp latent + DDPM + guidance + cross view attn.}, it is evident that semantic consistency between adjacent frames is significantly enhanced after incorporating guidance, as indicated by the noticeable improvements in both PSNR and SSIM scores in Table~1 of the main text. To further improve the cross-view consistency of high-frequency details, we have employed high-pass filtering. This approach aids in preserving the high-frequency details of the current frame, thus enhancing the semantic consistency of high-frequency details between consecutive frames. For instance, the enhanced cross-view consistency, particularly of the house on the left side in \cref{fig:ablation-full}, illustrates the effectiveness of adding the proposed modules.

Then, we conduct more detailed ablation studies on each proposed module in a Q\&A manner.

\paragraph{Q1:} Does DDIM inversion limit the reconstruction fidelity?

\begin{figure}[htbp] 
\centering 
\includegraphics[width=0.99\textwidth]{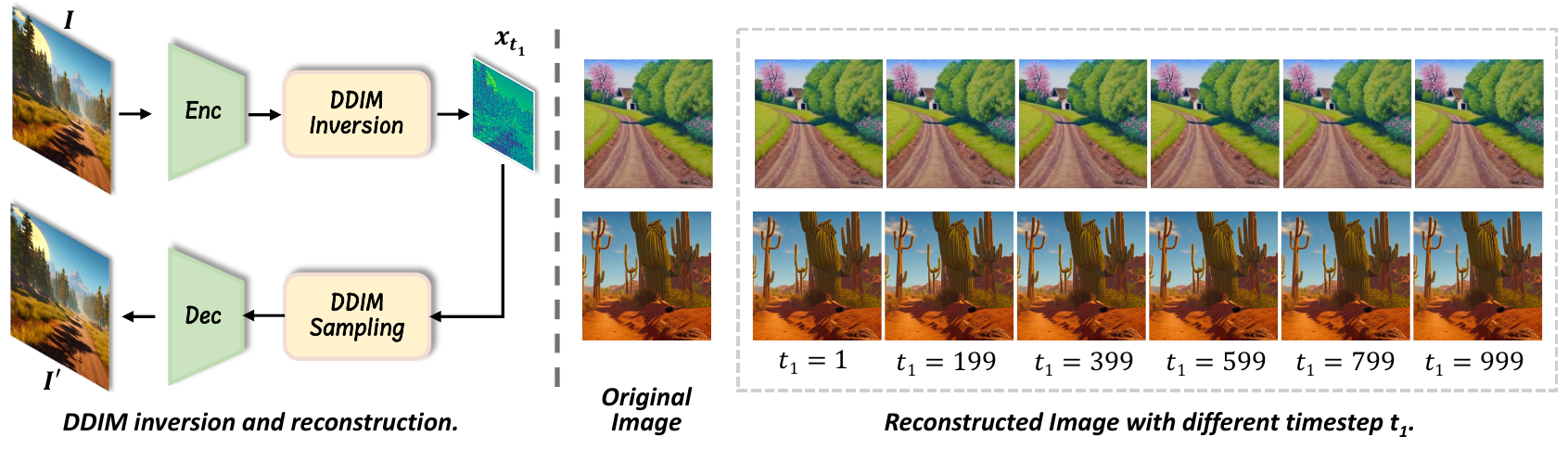} 
\vspace{-0.3cm}
\captionsetup{font=footnotesize}
\caption{\textbf{DDIM inversion and image reconstruction.} We illustrate the pipeline for evaluating the reconstruction performance using DDIM inversion. The left side is the pipeline and the right side is the reconstructed results at different timesteps $t_1$. We visualize two different result samples.}
\label{fig:ddim}
\end{figure}

Before warping latent, the most important thing is to ensure we can reconstruct the original image without any editing of the intermediate latent code. To this end, we establish a simple experiment. We obtain the intermediate latent code $x_{t_1}$ at different timestep $t_1$, denoising the noise, and decode the reconstructed image. In ~\cref{fig:ddim}, the left side is the pipeline of this experiment and the right side is the reconstructed images at different timestep $t_1$. We take two sample images as examples. From the results, we can figure out that the images can be reconstructed at different timestep $t_1$.

Based on the former discussion, regarding the top branch in Fig.~2 in the main text, the image at the current view can be reconstructed. The reconstruction of the top branch is the foundation of the feature-correspondence guidance and cross-view attention.

\paragraph{Q2:} Why do the image sequences become blurrier when generating more images, no matter whether warping the image or warping the latent code?

The $1^{st}$ and $2^{nd}$ rows in Fig.~3 in the main text show that the images become blurrier when generating more images. Besides, the reconstruction results in ~\cref{fig:ddim} show that DDIM inversion can reconstruct original images if there is no editing for the intermediate latent code, \textit{i.e.}, warping. It is straightforward that when we fly through, in other words, zoom in, the images, the images will become much blurrier. This is because the warping operation leads to non-integer pixel coordinates. Previous SOTA methods~\cite{liu2021infinite,li2022infinitenature,cai2023diffdreamer} train a refiner to add the details and inpainting or outpainting the missing region when the camera is moving. In this paper, we serve the pre-trained text-to-image diffusion model as a `refiner' due to its powerful generation capacity.

\paragraph{Q3:} Why DDPM forward is needed? 
% 过多的ddpm会使得自由度过大

\begin{figure}[htbp] 
\centering 
\includegraphics[width=0.6\textwidth]{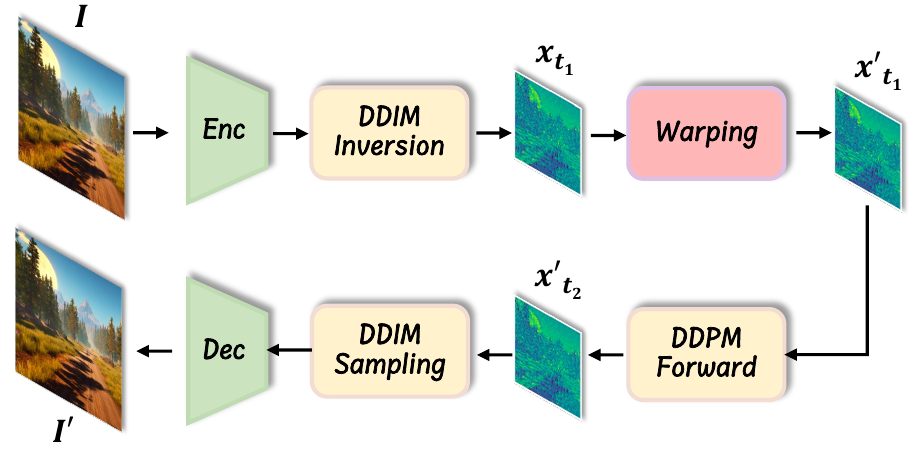} 
\vspace{-0.3cm}
\captionsetup{font=footnotesize}
\caption{\textbf{Ablation studies for DDPM forward without high-pass filtering.} To evaluate the necessity of DDPM forward operation, we conduct ablation experiments based on the simple pipeline. The corresponding ablation results are shown in \cref{fig:ddpm}}
\label{fig:ddpm-pipeline}
\end{figure}

Now we analyze the necessity of the DDPM forward process. As illustrated in Fig.~2 in the main text, we further apply DDPM forward after warping the latent code. Comparing the $5^{th}$ and $6^{th}$ rows in \cref{fig:ablation-full}, we can figure out that the image quality improves a lot after DDPM is applied. The details are enhanced and there is no distortion. The side effect of DDPM forward is that the correlation between adjacent views degrades because more degrees of freedom are introduced by DDPM.

\begin{figure}[htbp] 
\centering 
\includegraphics[width=0.99\textwidth]{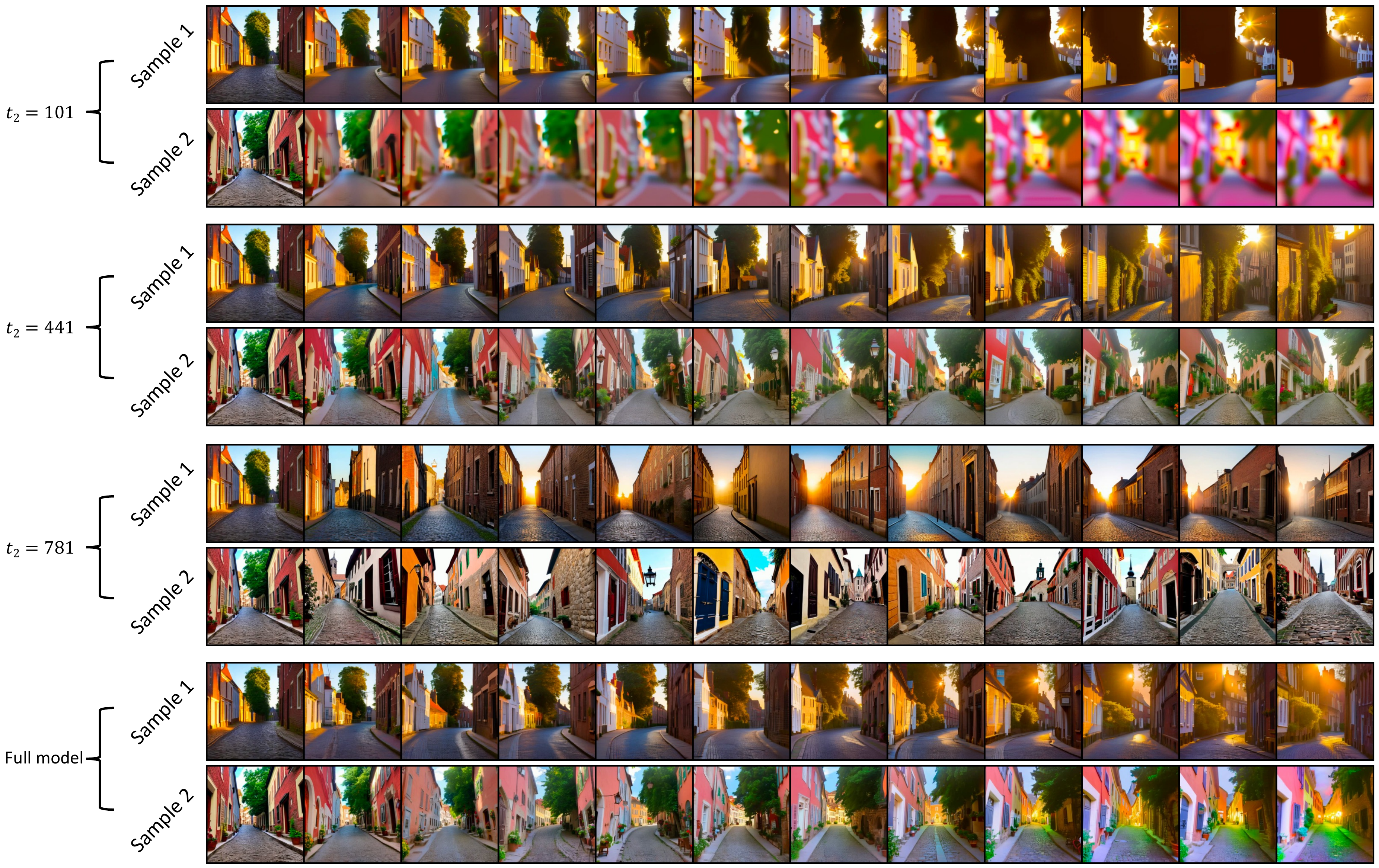} 
\vspace{-0.3cm}
\captionsetup{font=footnotesize}
\caption{\textbf{The visualization results for evaluating DDPM forward without high-pass filtering.} We illustrate the results based on the pipeline shown in \cref{fig:ddpm-pipeline}. We fix the warping timestep $t_1$ and generate image sequences with different DDPM forward timesteps $t_2$. To facilitate the comparisons, we further illustrate the generation results using our overall pipeline at last. We visualize every five frames per sample.}
\label{fig:ddpm}
\end{figure}

To evaluate how DDPM affects the generation results, we fix the warping timestep $t_1$ and illustrate the generated image sequences with DDPM at various timesteps $t_2$. The pipeline of this ablation experiment is shown in \cref{fig:ddpm-pipeline}. The results are shown in \cref{fig:ddpm}. A smaller $t_2$ means less degree of freedom for the diffusion model, which can result in blurring and distortion. As shown in the first two rows in \cref{fig:ddpm}, the generated images become blurrier when generating more images. A proper $t_2$ makes the geometry between adjacent views more consistent. In the $3^{rd}$ and $4^{rd}$ rows in \cref{fig:ddpm}, the geometry layout in the image sequences becomes consistent. For instance, the geometry of the house on the left side of the image looks roughly consistent. Moreover, the image quality is satisfied. As $t_2$ becomes larger, more random noise are added to the warped latent code. Though the image quality is promising, the consistency degrades a lot. As shown in the $5^{th}$ and $6^{th}$ rows, the consistency across adjacent views is much worse than $3^{th}$ and $4^{th}$ rows.

\begin{figure}[htbp] 
\centering 
\includegraphics[width=0.6\textwidth]{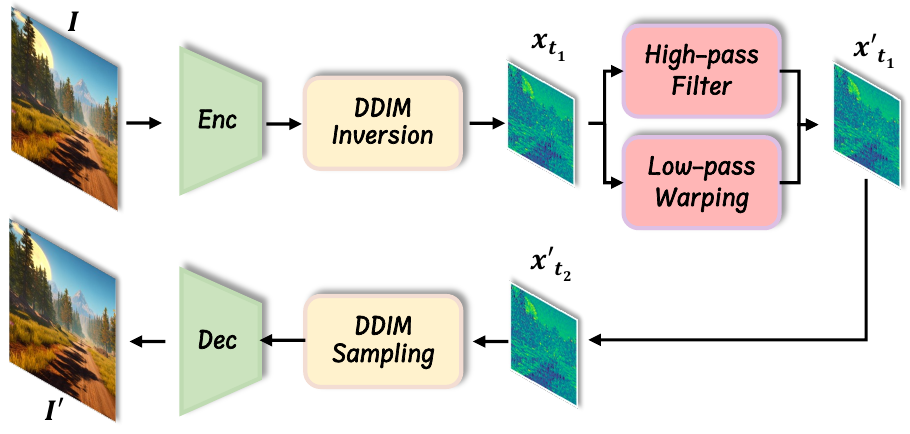} 
\vspace{-0.3cm}
\captionsetup{font=footnotesize}
\caption{\textbf{Ablation studies for high-pass filtering without DDPM forward.} To analyze if the DDPM can be replaced by the high-pass filter, we conduct a simple ablation based on this pipeline. The corresponding ablation results are shown in \cref{fig:ddpm-filter}.}
\label{fig:ddpm-filter-pipeline}
\end{figure}

This ablation demonstrates that the DDPM forward module with a proper timestep $t_2$ improves the image quality. But the consistency is still not satisfied. That's the reason why we further propose the feature-correspondence guidance strategy.

\begin{figure}[htbp] 
\centering 
\includegraphics[width=0.99\textwidth]{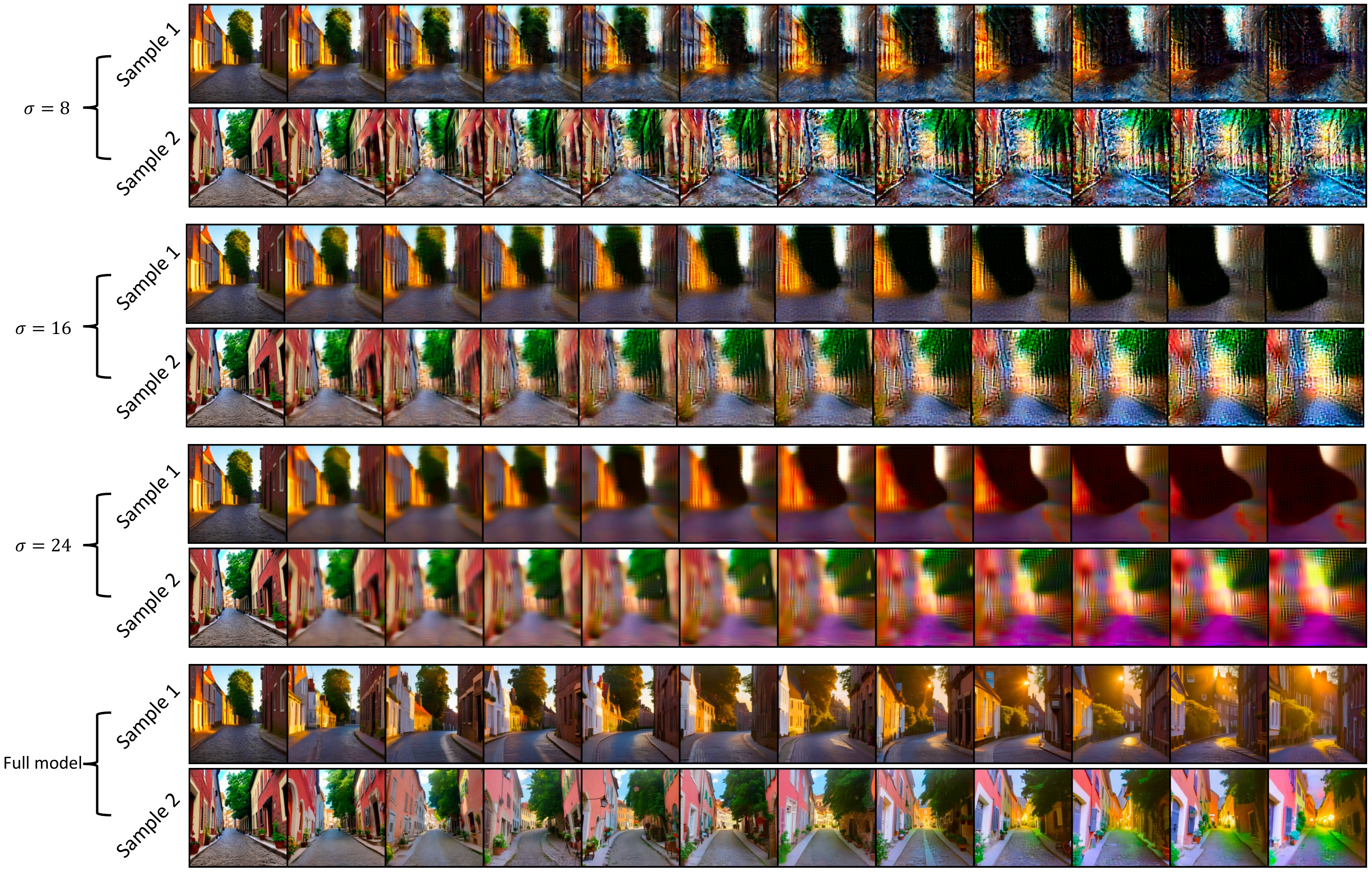} 
\vspace{-0.3cm}
\captionsetup{font=footnotesize}
\caption{\textbf{The visualization results for high-pass filter without DDPM forward.} To evaluate if we can preserve high quality from the previous view using the high-pass filter, we remove the DDPM forward module and visualize the image sequences with different threshold $\sigma$. To facilitate the comparisons, we further illustrate the generation results using our overall pipeline at last. We visualize every five frames per sample.}
\label{fig:ddpm-filter}
\end{figure}

\begin{figure}[htbp] 
\centering 
\includegraphics[width=0.99\textwidth]{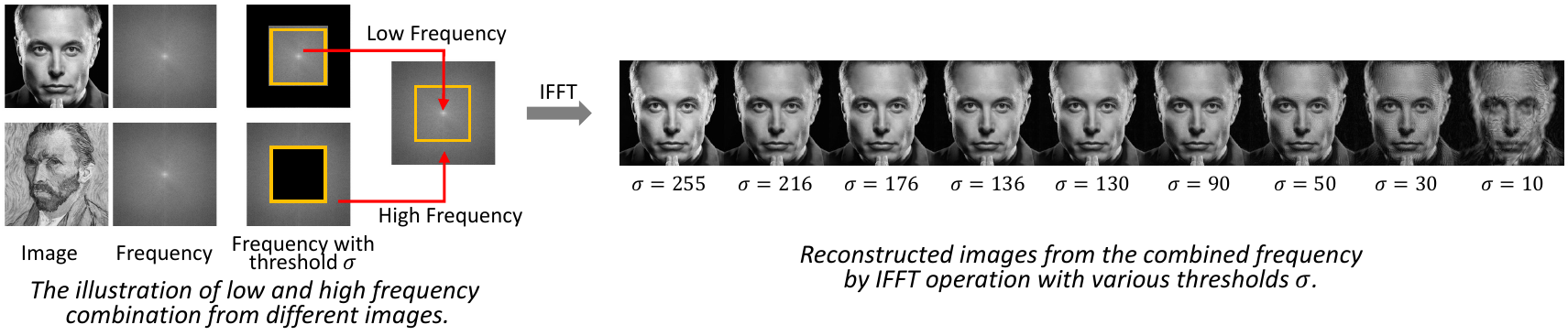} 
\vspace{-0.3cm}
\captionsetup{font=footnotesize}
\caption{\textbf{Toy experiments of low and high-frequency combination from different images.} Our toy experiments are illustrated on the left side. We combine the low frequency of Elon Musk and the high frequency of Vincent van Gogh's self-portrait with various threshold $\sigma$. The higher $\sigma$, the more low-frequency of Elon Musk is used. The results with various $\sigma$ are illustrated on the right side. Please zoom in for comparisons.}
\label{fig:fft}
\end{figure}

Moreover, since the high-pass filter preserves details from the previous view, is it possible to remove DDPM forward and only use the high-pass filter? To this end, we further conduct an ablation experiment. The pipeline for this ablation is shown in \cref{fig:ddpm-filter-pipeline}. In this pipeline, we remove DDPM forward operation and add the high-pass filter when warping the latent code. The experimental results are shown in \cref{fig:ddpm-filter}. We show two results for each $\sigma$. No matter how large the threshold $\sigma$ is, the high-pass filter cannot help to preserve the details from previous view. The reason is that the high-pass filter can only preserve high-frequency details from the previous view, rather than the low-frequency content. Combined with the results shown in \cref{fig:fft}, the low-frequency information dominants the content when combining frequencies from different images. As discussed before, the content would become much blurrier when warping the latent code, which motivates us to propose the feature-correspondence guidance strategy.

\paragraph{Q4:} Will the combination of low and high frequencies from different images break up the correlation of the frequency of the original image and introduce more errors?

To evaluate the feasibility of the frequency combination, we conduct a toy experiment, which is shown in \cref{fig:fft}. In this experiment, we first obtain the frequency from two different images and combine the frequencies given different threshold $\sigma$. As shown on the right side in \cref{fig:fft}, the content of Elon Musk does not change much with different $\sigma$, which demonstrates the feasibility of frequency combination. An extremely small $\sigma$, for instance, $\sigma = 10$, introduces excessive details from van Gogh's portrait.

\begin{figure}[htbp] 
\centering 
\includegraphics[width=0.99\textwidth]{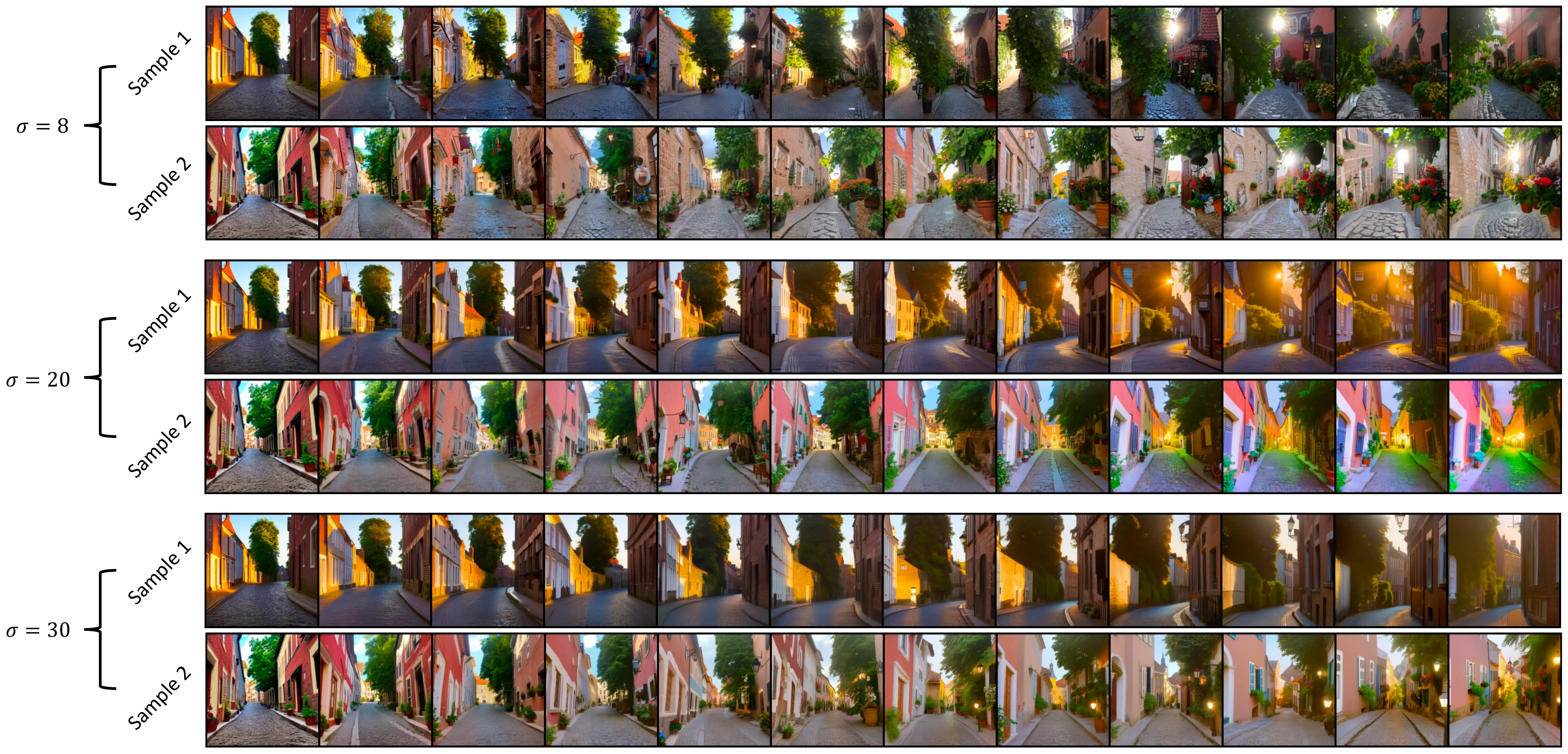} 
\vspace{-0.3cm}
\captionsetup{font=footnotesize}
\caption{\textbf{Ablation studies on high-pass filter.} We apply all the proposed modules with various thresholds $\sigma$ of the high-pass filter.}
\label{fig:q4}
\end{figure}

In this toy experiment, the content of the two images is extremely different. However, regarding the perpetual view generation task, the content of the adjacent view would not be so different. Now we analyze how different $\sigma$ affects the generation results. We apply all the proposed modules in this experiment and change the $\sigma$ value. The comparison results are shown in \cref{fig:q4}. As shown in \cref{fig:q4}, a small $\sigma$ neglects more low-frequency content from the previous view, which results in an inconsistency between images. A large $\sigma$ introduces less high-frequency details from the previous view. Though looks consistent, the generated images look not as realistic as $\sigma=20$.

\section{Additional qualitative comparisons}
\label{supp:quan}

\begin{figure*}[htbp] 
\centering 
\includegraphics[width=0.99\textwidth]{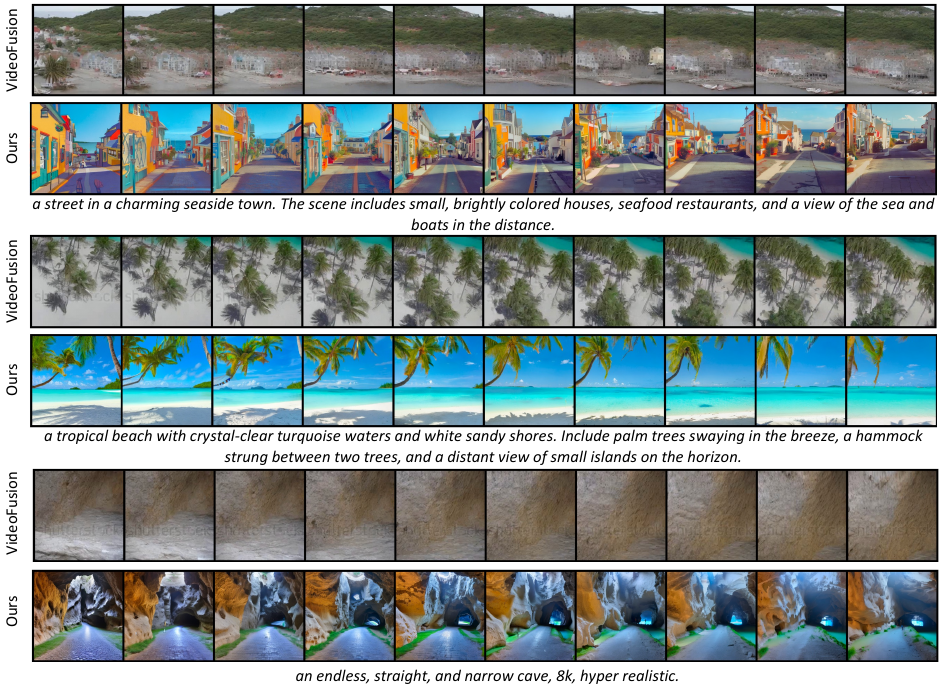} 
\caption{\textbf{Qualitative comparisons of VideoFusion~\cite{luo2023videofusion} and our DreamDrone}. We show the visualization results given three prompts and illustrate every five frames for each sample.}
\label{vid}
\end{figure*}

% 第二组：两个有监督一个无监督视频的对比
% text to video generation与我们的任务类似，最大的区别是text-to-video generation的任务无法受控于相机的位姿，且随着生成帧数的增加，生成质量会大打折扣。VideoFusion是视频生成任务SOTA的方法之一，我们提供了三组与其可视化对比的结果图。从图中可以看出，随着帧数的增加，VideoFUsion的生成结果变得模糊，且相机移动的效果没那么明显。我们的方法能够在生成高质量的连续场景的同时，保证帧与帧之间的几何一致性，能够明显的感受到相机在向前移动。山洞这种狭窄受限的场景生成更具有挑战性，VideoFusion在该prompt下效果不是很好，而我们的方法能够很好的表现出镜头在往前推进的效果。
Our task bears similarities to text-to-video generation, with the key difference being that text-to-video generation cannot be controlled by camera pose, and the quality significantly diminishes as the number of generated frames increases. VideoFusion~\cite{luo2023videofusion}, one of the state-of-the-art methods for video generation tasks, has been visually compared with our method, which is illustrated in \cref{vid}. It is evident that VideoFusion's generated results become blurry with an increase in frame count, and the effect of camera movement is less pronounced. In contrast, our method not only generates high-quality continuous scenes but also ensures geometric consistency between frames, clearly conveying the camera's forward movement. Generating scenes in constrained environments like caves is more challenging. VideoFusion does not perform well under such prompts, whereas our method effectively demonstrates the effect of the camera advancing forward.

\section{More visualization results}
\label{supp:vis}

\begin{figure*}[htbp] 
\centering 
\includegraphics[width=0.95\textwidth]{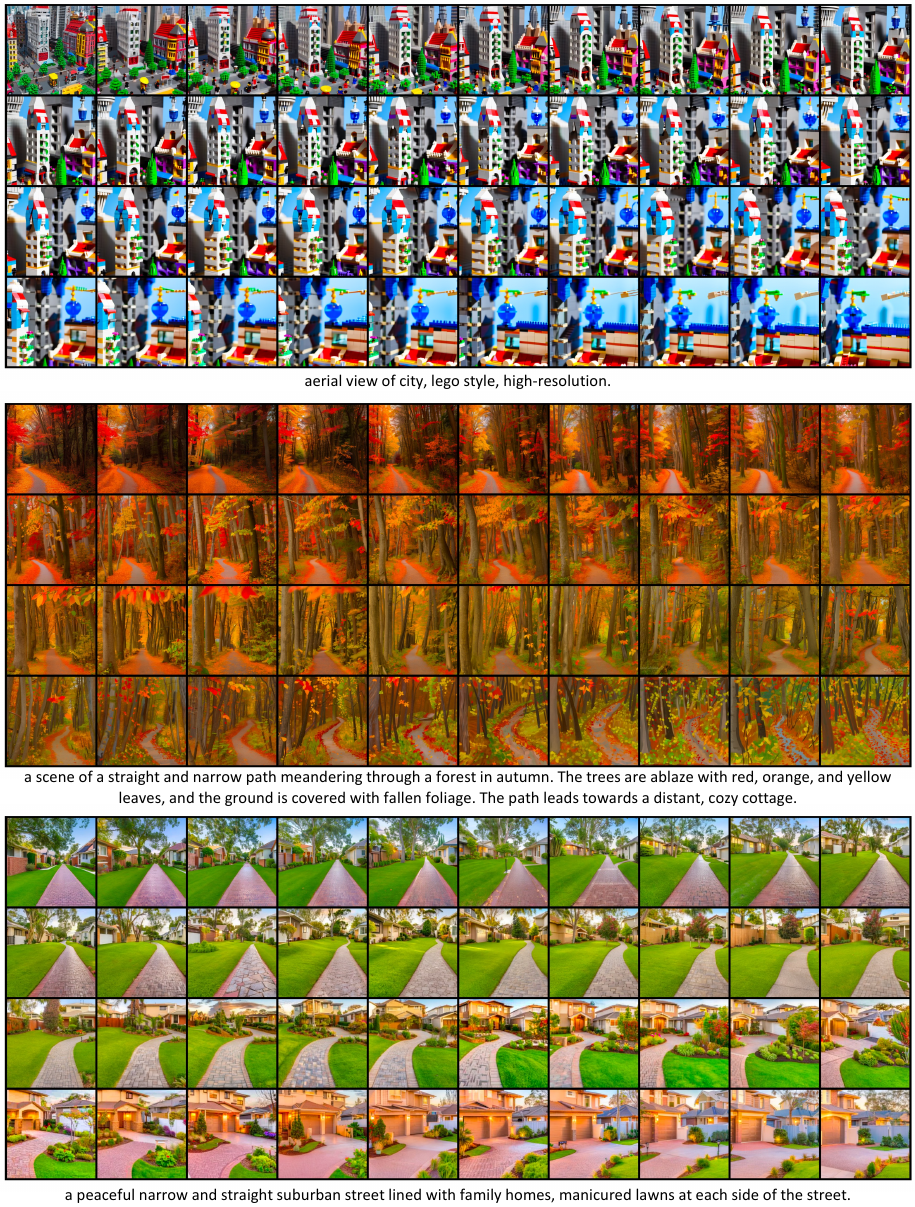} 
\caption{\textbf{Visualization results of our DreamDrone}. We generated 120 image sequences for each text prompt and visualized one image from every third frame to demonstrate the model's capability in producing diverse and stable visual outputs over time.}
\label{supp-vis}
\end{figure*}

\begin{figure*}[htbp] 
\centering 
\includegraphics[width=0.95\textwidth]{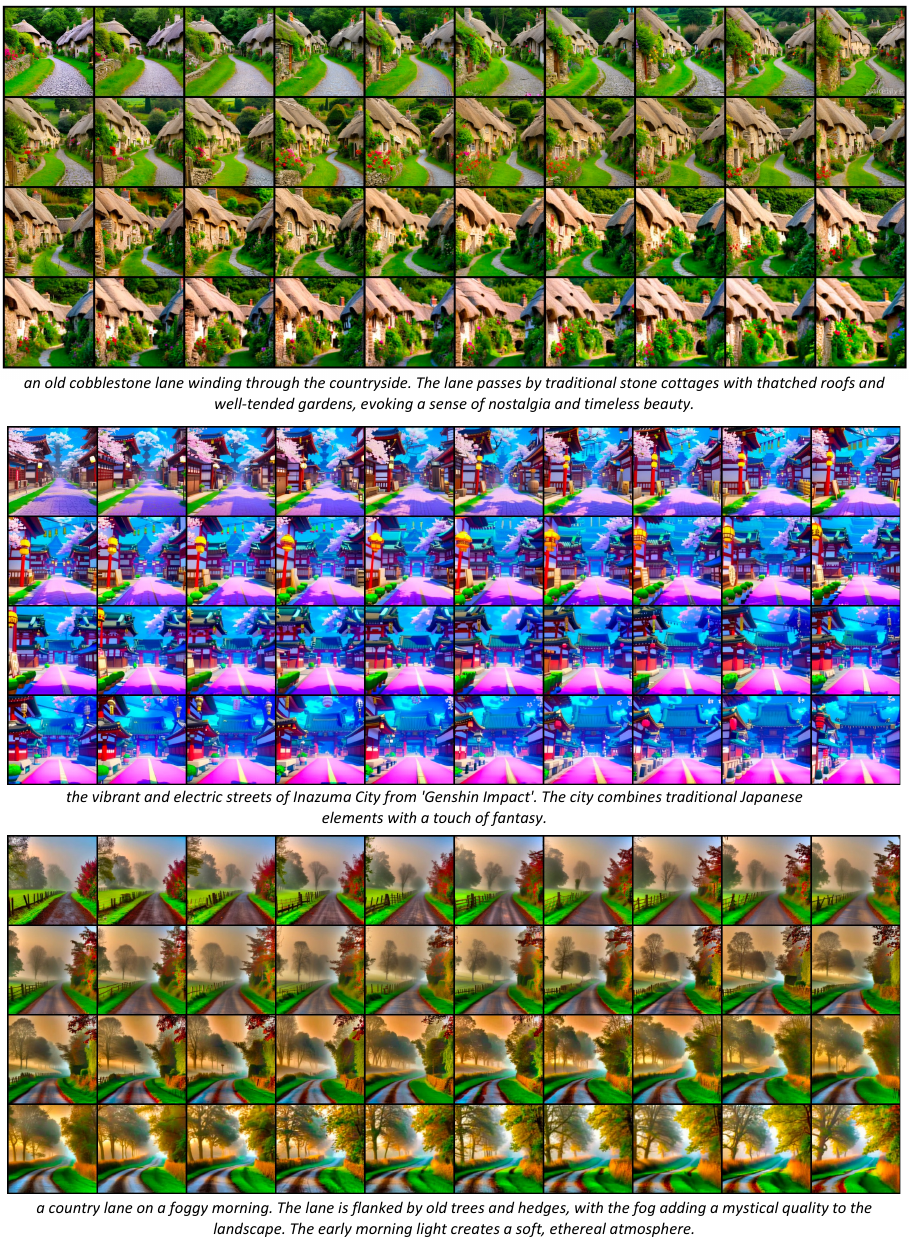} 
\caption{\textbf{Visualization results of our DreamDrone}. We generated 120 image sequences for each text prompt and visualized one image from every third frame to demonstrate the model's capability in producing diverse and stable visual outputs over time.}
\label{supp-vis1}
\end{figure*}

In this section, we provide more visualization results. We generate 120 images for each prompt and visualize one image from every third frame. Please refer to \cref{supp-vis,supp-vis1} for details.

\section{Limitation}

% 给定一个prompt，我们的方法能够在不需要任何训练或fine-tuning的情况下无限的对场景进行延伸，尽管如此，我们的方法还存在着一些limitation。首先，由于我们的方法是zero-shot、training-free的方法，故即使我们引入和feature-correspondence guidance和cross-frame self-attention module，相邻帧之间的高频细节对应的还不是非常完美（请参考附在补充材料里的视频材料）。其次，我们的方法非常依赖depth estimation的精度，虽然stable diffusion模型有一定的鲁棒性，但对于一些特殊style的场景，由于深度完全错误，导致生成的效果不尽如人意。我们会在后续的工作中改进如上缺点。

Given a prompt, our method can infinitely extend a scene without any training or fine-tuning. However, there are some limitations to our approach. Firstly, as our method is zero-shot and training-free, even with the introduction of feature-correspondence guidance and cross-frame self-attention modules, the correspondence of high-frequency details between adjacent frames is not yet perfect. Secondly, our method heavily relies on the accuracy of depth estimation. Although the stable diffusion model exhibits some robustness, for scenes with special styles, the entirely incorrect depth information leads to unsatisfactory generation results. We plan to address these shortcomings in our future work.

\section{Social impact}

We introduce a new method for creating perpetual scenes from text descriptions, making it easier for people to generate high-quality images without needing complex training or data. This breakthrough can help in various areas, such as making educational content more engaging, aiding in environmental planning, and giving creative professionals new tools to express their ideas. As this technology becomes available, it's important to use it wisely, ensuring it benefits society and does not contribute to misinformation or unethical use. In essence, DreamDrone offers exciting possibilities for innovation while emphasizing the need for responsible use.